\documentclass[final]{cvpr}

\usepackage{times}
\usepackage{epsfig}
\usepackage{graphicx}
\usepackage{amsmath}
\usepackage{amssymb}
\usepackage{mathtools}
\usepackage{subfig}
\mathtoolsset{showonlyrefs=true}
\usepackage{bbm}
\usepackage{bm}
\usepackage{multirow}
\usepackage{booktabs}

\usepackage[pagebackref=true,breaklinks=true,letterpaper=true,colorlinks,bookmarks=false]{hyperref}

\newtheorem{problem}{Problem}

\newtheorem{newexample}{Example}

\newcommand{\argmax}{\mathop{\rm argmax}\limits}

\begin{document}

\title{Generalized Domain Adaptation}
\author{Yu Mitsuzumi~~~Go Irie~~~Daiki Ikami~~~Takashi Shibata\\
NTT Communication Science Laboratories, NTT Corporation, Japan\\
{\tt\small \{yu.mitsuzumi.ae, daiki.ikami.ef\}@hco.ntt.co.jp, \{goirie, t.shibata\}@ieee.org}
}

\maketitle

\begin{abstract}
Many variants of unsupervised domain adaptation (UDA) problems have been proposed and solved individually. Its side effect is that a method that works for one variant is often ineffective for or not even applicable to another, which has prevented practical applications. 
In this paper, we give a general representation of UDA problems, named Generalized Domain Adaptation (GDA). GDA covers the major variants as special cases, which allows us to organize them in a comprehensive framework.
Moreover, this generalization leads to a new challenging setting where existing methods fail, such as when domain labels are unknown, and class labels are only partially given to each domain.
We propose a novel approach to the new setting. The key to our approach is self-supervised class-destructive learning, which enables the learning of class-invariant representations and domain-adversarial classifiers without using any domain labels.
Extensive experiments using three benchmark datasets demonstrate that our method outperforms the state-of-the-art UDA methods in the new setting and that it is competitive in existing UDA variations as well.
\end{abstract}

\section{Introduction}\label{sec:intro}

\begin{figure}
    \centering
    \includegraphics[width=\linewidth]{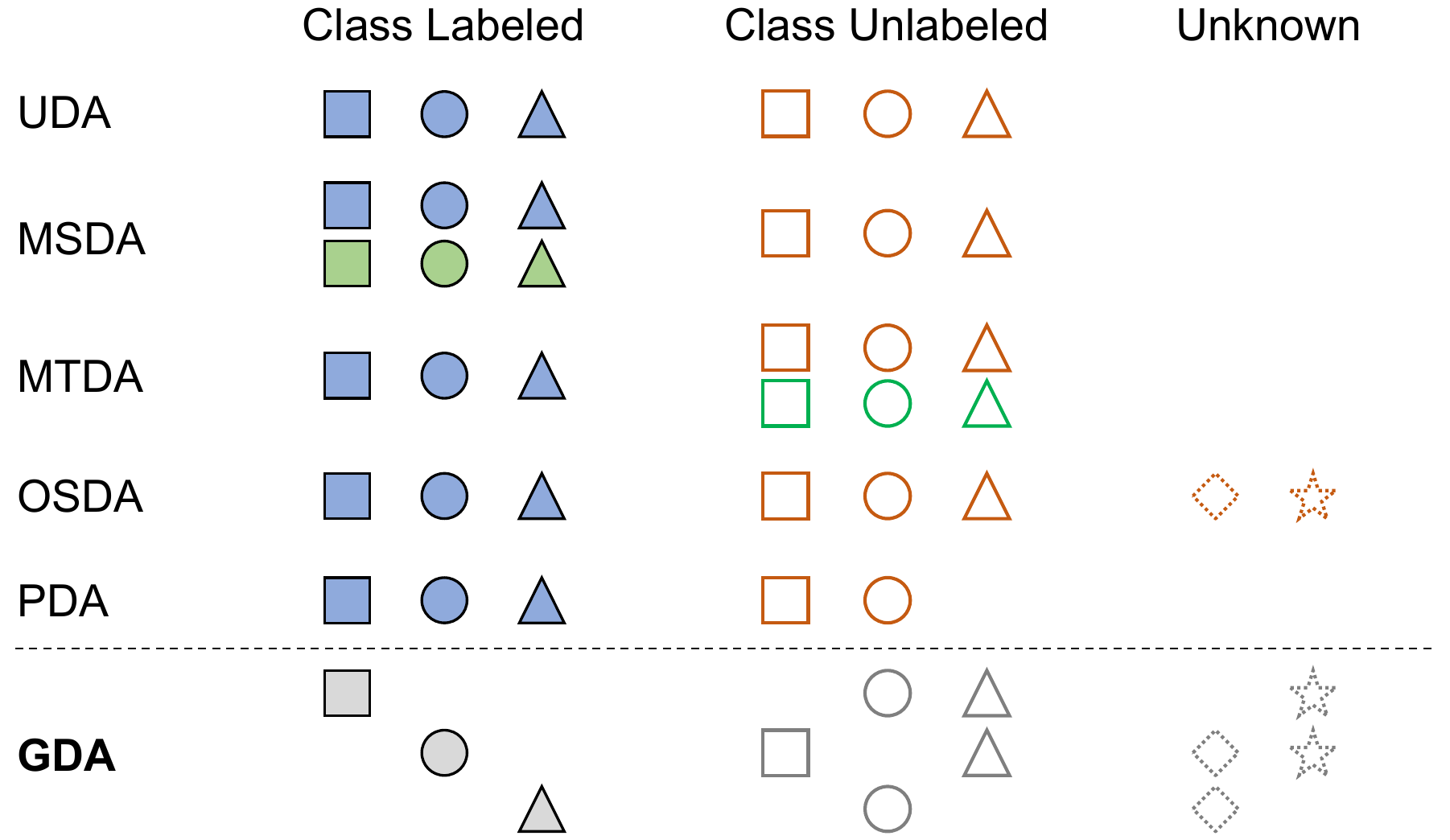}
    \vspace{-13pt}
    \caption{{\bf Schematic overview of Generalized Domain Adaptation (GDA)}. GDA covers major existing UDA problems as its special cases by imposing some constraints on classes and/or domains (represented in the shapes and colors of symbols, respectively). Moreover, it provides new challenging settings where domain labels are unknown, and class labels are given to only a subset in each domain.
    }
    \label{fig:generalized_domain_adaptation}
    \vspace{-12pt}
\end{figure}

Deep learning is data-hungry. It performs remarkably well in a domain that has a sufficient amount of labeled data, but its performance suffers significantly in one that does not. Unsupervised domain adaptation (UDA) aims to resolve this problem by transferring a model learned for a label-rich source domain to a label-less target domain. 

Besides the standard UDA, which assumes a fully labeled source domain and a completely unlabeled target domain, a number of variants have been proposed to address more complex and practical problems. 
Major variants are illustrated in Fig.~\ref{fig:generalized_domain_adaptation}.  
One representative example is multi-source domain adaptation (MSDA) or multi-target domain adaptation (MTDA), an extension to the case where there is more than one source or target domain~\cite{zhao2018multisource, peng2019moment, gholami2020multitarget, gholami2020multitarget}. 
Open set domain adaptation (OSDA) and partial domain adaptation (PDA) address the case where the class sets of the source and target do not match, i.e., there exist unknown classes~\cite{panareda2017open, saito2018open, liu2019separate, bucci2020effectiveness, cao2018partial}. 
Some extensions of these variants have also been studied~\cite{mancini2018boosting,chen2019blending,you2019universal}.

Most of these variants have been proposed independently and solved individually.
A negative side of this history is that a method that works for one variant may not work for or even be applicable to another.
In reality, it is rarely possible to identify which variant of the problems one is facing, which requires a costly trial and error to figure out the type of problem or find a satisfactory solution.
Moreover, a real problem is often a combination of these variants, in which case none of the methods will eventually be applicable.

In this work, we aimed to overcome this problem. Our approach is to first consider giving a new general representation of the UDA problems that covers all these major UDA variants and then implement a method for solving them.
Definitions of the most existing UDA problems discussed above assume a clear distinction between the source and target domains, and whether or not class labels are available is determined on a domain-by-domain basis.
Instead, in our generalized representation, which we call Generalized Domain Adaptation (GDA), everything is determined on a sample-by-sample basis; each sample is given a class label, a domain label, and indexes indicating whether or not these labels are available. 
In Sec.~\ref{sec:GDA}, we will show that this slight difference in perspective allows us to represent all the major variants and their combinations as GDA's special cases.

Moreover, GDA brings a brand new challenge as illustrated in Fig.~\ref{fig:generalized_domain_adaptation}, which is not just a straightforward combination of the existing variants.
The key properties of this setting are that the domain labels are completely unknown for all the samples, and the class labels are given to only a subset of the classes of each domain.
Indeed, such a setting arises in practice, for example, for data coming from multiple institutions where the acquisition processes are unknown.
Nonetheless, it has not received much attention so far.
As shown later in our experiments (Sec.~\ref{sec:experiments}), the state-of-the-art UDA methods applicable to this setting, if forced, suffer from severe performance degradations.

We propose self-supervised class-destructive learning to accurately estimate the domain of each sample, which is essential missing ingredient to solve the new problem.
The assumptions behind our approach are that (1) class information of an image is strongly dependent on its local structural information (e.g., shape and part connection); and that (2) domain and class information are independent of each other. 
Based on these assumptions, our method first transforms an image into a ``class-indistinguishable" form by randomly shuffling the positions of its pixel blocks to break its local structure and then performs self-supervised learning to capture class-independent information.
This enables learning of class-invariant and domain-variant representations without using any domain labels, making it possible to train a domain-invariant classifier with a simple domain-adversarial learning approach. Our method works even for the case where the class labels are only partially available for each domain.
Furthermore, our method can readily cope with open set settings by integrating a joint label-network optimization framework~\cite{tanaka2018joint}.
Thorough experiments on three datasets demonstrate that our method outperforms the state-of-the-art methods in the new setting and is highly competitive in the existing UDA problems.

Our main contributions include: 
(1) a general representation of UDA problems named Generalized Domain Adaptation (GDA);
(2) new UDA settings where existing methods fail; and 
(3) a novel domain label estimation method based on self-supervised class-destructive learning.


\section{Related Work}\label{sec:related_works}
Given a fully labeled source domain and an unlabeled target domain, the task of (standard) UDA is to train a classifier for the target domain by bridging the gap between the two domains (domain shift). 
Numerous methods have been proposed~\cite{tzeng2014deep, long2015learning, long2016unsupervised, sun2016coral, sun2016deep, ganin2014unsupervised, ganin2016domain, tzeng2017adversarial}, but most of them are based on strong assumptions that may not be feasible in practice -- the number of domains is always two, which distribution each data sample comes from (i.e., domain labels) is known, and class sets of both domains are perfectly consistent.
This has led to a growing stream of various UDA variants and attempts to develop more realistic and flexible methodologies. The major variants are as follows:

\medskip
\noindent {\bf Multi-Source/Target UDA}.
Multi-source domain adaptation (MSDA)~\cite{mansour2009domain, sun2011twostage, zhao2018multisource, peng2019moment} and multi-target domain adaptation (MTDA)~\cite{yu2018multitarget, gholami2020multitarget} consider UDA where either the source or target domain consists of multiple subdomains.
While these studies assume that the subdomain labels of all samples are available, some others assumed that they are unavailable in the source~\cite{mancini2018boosting} or target domain (Blending-target domain adaptation: BTDA)~\cite{chen2019blending, pmlr-v97-peng19b}.

\medskip
\noindent {\bf Open Set and Partial UDA}.
These problems consider UDAs with ``unknown" classes.
Open set domain adaptation (OSDA) assumes that private classes exist in the target domain~\cite{saito2018open, liu2019separate, bucci2020effectiveness} or both~\cite{panareda2017open}. 
In partial domain adaptation (PDA)~\cite{cao2018partial, cao2019learning}, the class set of the source is a superset of that of the target. 
Universal domain adaptation (UniDA)~\cite{you2019universal} is an integration of OSDA and PDA. Weakly supervised OSDA has also been studied~\cite{tan2019weakly}. 
Multi-source open set domain adaptation (MS-OSDA) is a combination of MSDA and OSDA~\cite{rakshit2020mosda}.
A variant of MS-OSDA has also been investigated~\cite{xu2018deep}.

\begin{table*}[t]
	\begin{center}
		\caption{{\bf GDA representations of existing and new UDA problems}. Problems and corresponding constraints are listed.}
        \vspace{-9pt}
		\label{tab:existingUDA}
		\scalebox{0.91}{
			\begin{tabular}{l|lllll}
			\hline \hline
			 \multirow{2}{*}{Problem}& \multicolumn{1}{c}{\multirow{2}{*}{$d$}} &  \multicolumn{3}{c}{Class label set condition} & \multicolumn{1}{c}{\multirow{2}{*}{$\delta_d$}} \\ 
			 & & \multicolumn{1}{c}{Openness} & \multicolumn{1}{c}{Target domain} & \multicolumn{1}{c}{Source domain} & \\ \hline
             UDA & $d\in\{1,2\}$ & $\mathcal{C}_1=\mathcal{C}_2=\mathcal{C}$ & $\mathcal{L}_2 = \emptyset,~\mathcal{U}_2 = \mathcal{C}_2$ & $\mathcal{L}_1 = \mathcal{C}_1 = \mathcal{L},~\mathcal{U}_1 = \emptyset$ &  $\forall \delta_d = 1$ \\
             MSDA & $d \in \mathbb{N}$ & $\forall i,~\mathcal{C}_i=\mathcal{C}$ & $\exists! j,~\mathcal{L}_j = \emptyset,~\mathcal{U}_j = \mathcal{C}_j$ & $\forall i \neq j,~\mathcal{L}_i = \mathcal{C}_i =  \mathcal{L},~\mathcal{U}_i = \emptyset$ & $\forall \delta_d = 1$ \\	
             OSDA & $d\in\{1,2\}$ & $\mathcal{C}_1\subset\mathcal{C}_2=\mathcal{C}$ & $\mathcal{L}_2 = \emptyset,~\mathcal{U}_2 = \mathcal{C}_2$ & $\mathcal{L}_1 = \mathcal{C}_1 = \mathcal{L},~\mathcal{U}_1 = \emptyset$ &  $\forall \delta_d = 1$ \\
             MS-OSDA & $d \in \mathbb{N}$ & $\begin{aligned}\exists&! j, ~\mathcal{C}_j=C, \\ \forall &i \neq j, ~\mathcal{C}_i\subset\mathcal{C}_j\end{aligned}$ & $\exists! j,~\mathcal{L}_j = \emptyset,~\mathcal{U}_j = \mathcal{C}_j$ & $\forall i \neq j,~\mathcal{L}_i = \mathcal{C}_i =  \mathcal{L},~\mathcal{U}_i = \emptyset$ & $\forall \delta_d = 1$ \\
             BTDA & $d \in \mathbb{N}$ & $\forall i,~\mathcal{C}_i=\mathcal{C}$ & $\forall i \neq j,~\mathcal{L}_i = \emptyset, ~\mathcal{U}_i = \mathcal{C}_i$ & $\exists! j,~\mathcal{L}_j = \mathcal{C}_j = \mathcal{L},~\mathcal{U}_j = \emptyset$ & $\delta_d = 
                \begin{cases} 1 & (d = j) \\
                    0 & (d \neq j)
                \end{cases}$ \\ \hline
            GDA1 & $d \in \mathbb{N}$ & ($\mathcal{L}\subseteq\mathcal{C}$) & \multicolumn{2}{c}{$\exists~i,j,~\mathcal{L}_i \neq \mathcal{L}_j, ~\forall k,~\mathcal{L}_k \cap \mathcal{U}_k = \emptyset$} & $\forall \delta_d = 0$ \\
            GDA2 & $d \in \mathbb{N}$ & ($\mathcal{L}\subseteq\mathcal{C}$) & \multicolumn{2}{c}{$\exists~i,j,~\mathcal{L}_i \neq \mathcal{L}_j, ~\forall k,~\mathcal{L}_k \subset \mathcal{U}_k$} & $\forall \delta_d = 0$ \\
            \hline \hline
            \end{tabular}
		}
        \vspace{-20pt}
	\end{center}
\end{table*}

\medskip
\noindent In this paper, we propose GDA, a general representation that covers all these major UDA variants.
As described in the next section, GDA allows for a systematic discussion of the UDA variants that have been studied independently. It also reveals pivotal settings that have not been considered before and the missing parts essential to solve them.


\section{Generalized Domain Adaptation}\label{sec:GDA}
We first give the formal definition of our GDA.
\begin{problem}{\bf (GDA)} \label{prob:gda}
Suppose we are given a set of tuples 
\begin{align*}
\mathcal{D}=\{(\mathbf{x}, y, d, \delta_y, \delta_d)|\mathbf{x} \in \mathbb{R}^n, y, d \in \mathbb{N}, \delta_y, \delta_d \in \{0, 1\}\},
\end{align*}
where $\mathbf{x}$ is a sample (e.g., image) of $n$ dimensions, $y$ is a class label, $d$ is a domain label, and $\delta_y$ and $\delta_d$ are flags indicating whether the class and domain labels are available (1) or not (0) for training, respectively,

The task is to find a class classifier $F$ that satisfies
\begin{align*}
    F(\mathbf{x}) = 
    \begin{cases} y & (y \in \mathcal{L}) \\
    {\rm UNK} & (y \in \mathcal{C}\setminus\mathcal{L}),
    \end{cases}
\end{align*}
where {\rm UNK} is the symbol for unknown classes, and
\begin{align*}
    \mathcal{C} &= \left\{y | (\mathbf{x}, y, d, \delta_y, \delta_d) \in \mathcal{D}\right\}, \\
    \mathcal{L} &= \left\{y | (\mathbf{x}, y, d, \delta_y, \delta_d) \in \mathcal{D}, \delta_y = 1 \right\}.
\end{align*}
\end{problem}

\noindent Intuitively, $\mathcal{C}$ is the set of all classes contained in $\mathcal{D}$, and $\mathcal{L}$ is the ``known" (i.e., labeled) subset of $\mathcal{C}$. 
Given $\mathbf{x}$, $F$ is required to output its correct class label $y$ iff it is in $\mathcal{L}$, or to detect it as unknown otherwise.
In GDA, the availability of the class and domain label is controlled on a sample-by-sample basis. This makes it possible to encompass new cases, for example, where the class labels are only available for a part of the classes or samples within the same domain, or the domain labels are not available at all.

\subsection{Representing Existing UDA Problems} \label{sec:gda_existing}
We show that GDA (Problem~\ref{prob:gda}) turns into various existing UDA problems by imposing proper constraints. 
Representative examples are listed in Table~\ref{tab:existingUDA}.
For brevity, we use $\mathcal{C}_i$, $\mathcal{L}_i$ and $\mathcal{U}_i$ for the set of all classes, known classes, and unknown classes within domain $i$, respectively.
\begin{align*}
    \mathcal{C}_i &= \left\{y | (\mathbf{x}, y, d, \delta_y, \delta_d) \in \mathcal{D}, d = i\right\}, \\
    \mathcal{L}_i &= \left\{y | (\mathbf{x}, y, d, \delta_y, \delta_d) \in \mathcal{D}, d = i, \delta_y = 1 \right\}, \\
    \mathcal{U}_i &= \left\{y | (\mathbf{x}, y, d, \delta_y, \delta_d) \in \mathcal{D}, d = i, \delta_y = 0 \right\}.
\end{align*}

The standard UDA can be represented by limiting the number of domains to two ($d \in \{1,2\}$) sharing the same class set ($\mathcal{C}_1 = \mathcal{C}_2 = \mathcal{C}$), with the domain labels available for all the samples ($\forall \delta_d = 1$) and the class labels fully available for only those in one of the two domains ($\mathcal{L}_1 = \mathcal{C}_1 = \mathcal{L},~\mathcal{U}_1 = \emptyset,~\mathcal{L}_2 = \emptyset,~\mathcal{U}_2 = \mathcal{C}_2$). 
Discarding the condition on the number of domains ($d \in \mathbb{N}$) leads to multi-domain settings. For example, MSDA is represented by imposing constraints so that the class labels are unavailable in a certain domain ($\exists! j,~\mathcal{C}_j = \mathcal{C},~\mathcal{L}_j = \emptyset$) but fully accessible in the others ($\forall i \neq j,~\mathcal{C}_i = \mathcal{C},~\mathcal{L}_i = \mathcal{C}_i = \mathcal{L},~\mathcal{U}_j = \emptyset$).
OSDA~\cite{saito2018open, liu2019separate, bucci2020effectiveness} contains unknown classes in the target domain ($\mathcal{C}_1\subset\mathcal{C}_2=\mathcal{C}$).
PDA~\cite{cao2018partial,cao2019learning} and UniDA~\cite{you2019universal} can be written in the same way. 
Combinations and extensions of these problems, such as MS-OSDA~\cite{rakshit2020mosda} and BTDA~\cite{chen2019blending, pmlr-v97-peng19b}, can also be represented as in Table~\ref{tab:existingUDA}.

\begin{figure*}
    \centering
    \includegraphics[width=\linewidth]{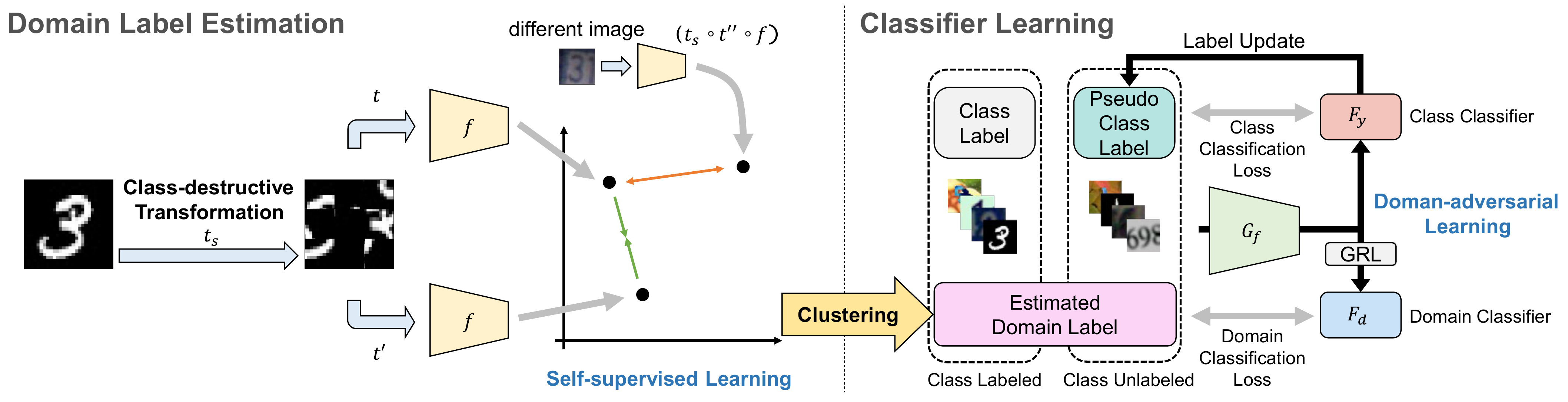}
    \vspace{-17pt}
    \caption{{\bf Overview of our approach to GDA}. We first apply class-destructive transformation to images of unknown domains and then perform self-supervised learning on them to estimate their domain labels. We then perform domain-adversarial learning with the estimated domain labels to learn a domain-invariant classifier.}
    \label{fig:method}
    \vspace{-8pt}
\end{figure*}

\subsection{Deriving New UDA Problems} \label{sec:gda_new}
The systematic representations in Table~\ref{tab:existingUDA} reveal that most of the existing UDA problems impose similar constraints of only limited types, which suggests a great potential to open up new UDA problems.
In particular, all the UDA problems assume that the domain labels are visible ($\forall \delta_d = 1$) for at least one domain, which naturally reminds us to consider a ``blind" domain adaptation problem where no domain labels are available for any domain, i.e., $\forall \delta_d = 0$.
This is extremely challenging, especially when the labeled classes are inconsistent between the domains, i.e., $\exists~i,j,~\mathcal{L}_i \neq \mathcal{L}_j$, because no explicit information for disentangling the class and domain features is given.

Depending on whether the class labels are fully assigned to samples within the same class (i.e., $\forall k,~\mathcal{L}_k \cap \mathcal{U}_k = \emptyset$) or only given to some of them (i.e., $\forall k,~\mathcal{L}_k \subset \mathcal{U}_k$), we can define the following two versions of GDA:
\begin{newexample} {\bf (GDA1)} 
\begin{align*}
    & \forall \delta_d = 0,\\
    & \exists~i,j,~\mathcal{L}_i \neq \mathcal{L}_j,~\forall k,~\mathcal{L}_k \cap \mathcal{U}_k = \emptyset.
\end{align*}
\end{newexample}
\begin{newexample}
{\bf (GDA2)} 
\begin{align*}
    & \forall \delta_d = 0,\\
    & \exists~i,j,~\mathcal{L}_i \neq \mathcal{L}_j,~\forall k,~\mathcal{L}_k \subset \mathcal{U}_k.
\end{align*}
\end{newexample}
Note that these problems do not make any assumptions on ``openness"; they can be either a case where all the classes are known ($L=C$) or some of them are unknown ($L \subset C$).

None of the existing methods are directly applicable to these problems, as they all assume that the domain labels are known for at least one domain. They can be applied, if forced, by considering the labeled and unlabeled data (across multiple domains) as the ``source" and ``target", respectively. However, as shown later in Sec.~\ref{sec:experiments}, the performance is far from satisfactory.


\section{Method}\label{sec:method}
We propose an approach to solving the new problems.
An overview is shown in Fig.~\ref{fig:method}.
Our method comprises two major steps: estimating the domain labels for all samples and learning a domain-invariant classifier using the estimated domain labels.
 
\subsection{Domain Label Estimation} \label{sec:domain_label_estimation}
One primary source of the difficulty of these problems is that the domain labels are unknown for all images.
The first step, which is the core of our proposed method, estimates the domain labels in a self-supervised learning manner.

\medskip
\noindent {\bf Self-supervised Class-destructive Learning}.
A straightforward approach to estimating the domain labels would be to assume that images belonging to the same domain form a tight cluster in their feature space and to estimate their domain labels by feature clustering.
In fact, an existing approach to BTDA~\cite{chen2019blending} is based on a similar assumption to estimate the (sub)domain labels hidden in the target domain.
Unfortunately, the results are not as expected; the classes and domains are easily confused in the new problems where the class labels are only given to a subset of images, resulting in the formation of class-dependent clusters.

\begin{figure}
    \centering
    \includegraphics[width=\linewidth]{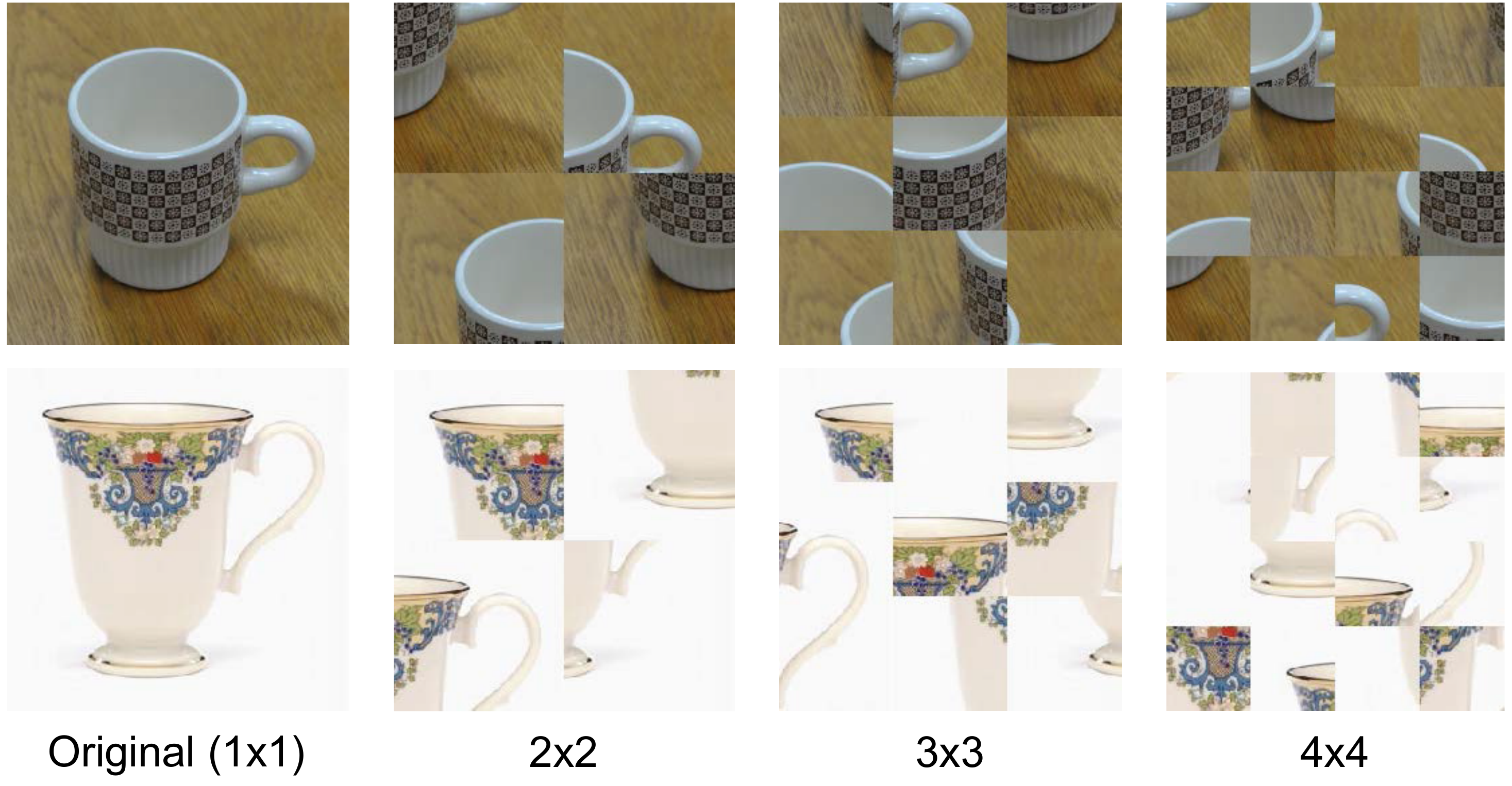}
    \vspace{-15pt}
    \caption{{\bf Examples of transformed images}. Two ``mug" images of different domains in Office-31. The object class is harder to recognize as the number of grid partitions is increased, but their domains are still identifiable.}
    \label{fig:shuffled_images}
    \vspace{-8pt}
\end{figure}

We aim to prevent the formation of such class-dependent clusters by applying image transformation that aggressively disrupts the class-dependent information of images and using the transformed images for self-supervised feature learning.
Specifically, we focus on local structural information of images which is an important clue representing the class of images, such as the shape of an object or the connections between its parts, and destroy it by using a transformation that divides the original image into several pixel blocks and randomly shuffles their positions.
Fig.~\ref{fig:shuffled_images} shows examples of transformed images at different pixel block sizes.
When the partition is coarse, the shape of the object and its parts are still recognizable. However, as the number of partitions is increased, even the parts are fragmented and become unrecognizable; meanwhile, it is relatively easy to identify which domain each image comes from owing to the remaining global information of the pixels.

\begin{figure}[t]
\vspace{-15pt}
\centering
\subfloat[][Domain NMI]{\includegraphics[clip, width=.5\columnwidth]{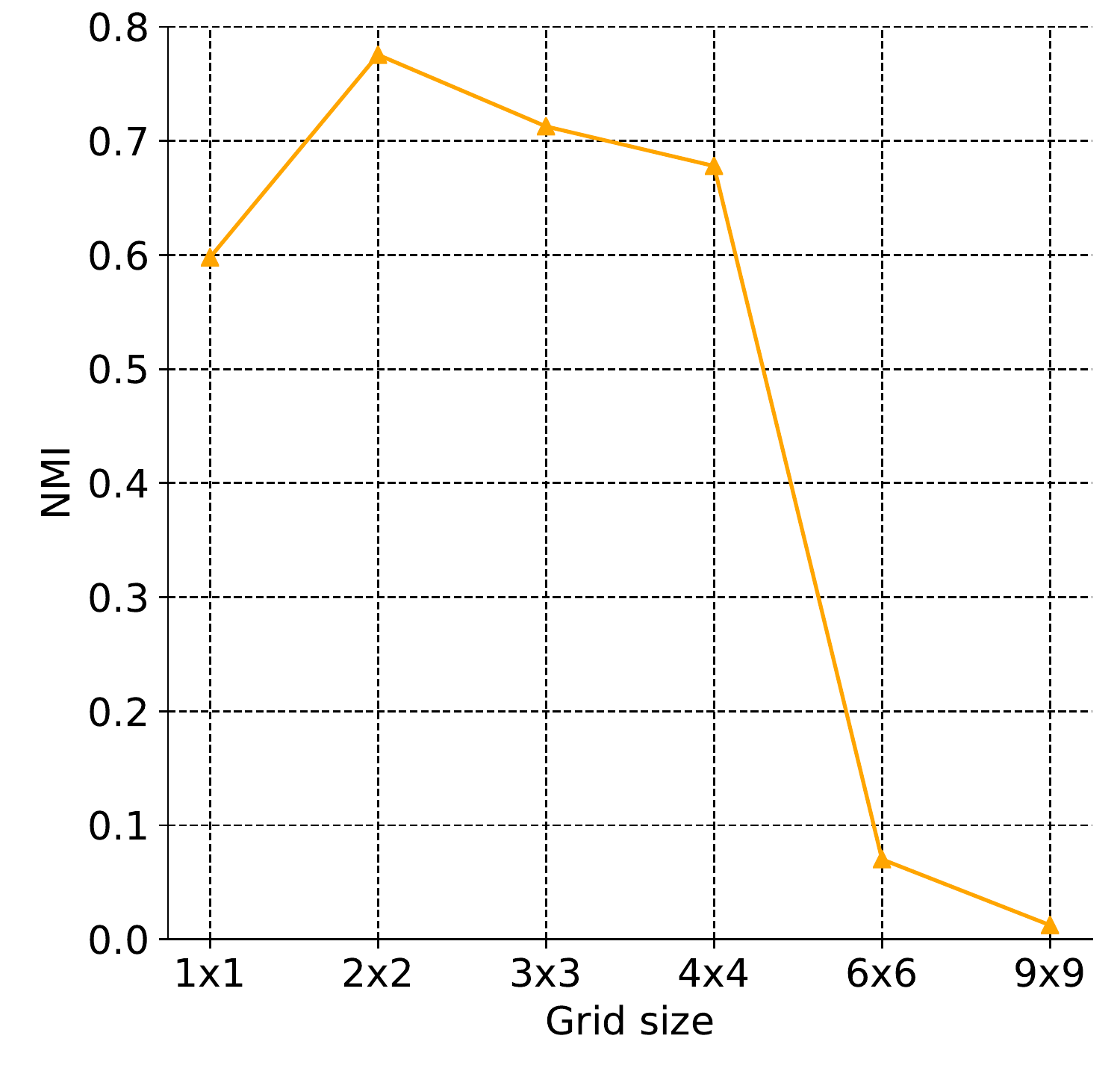}
\label{fig:nmi_analysis-b}}
\subfloat[][Class NMI]{\includegraphics[clip, width=.5\columnwidth]{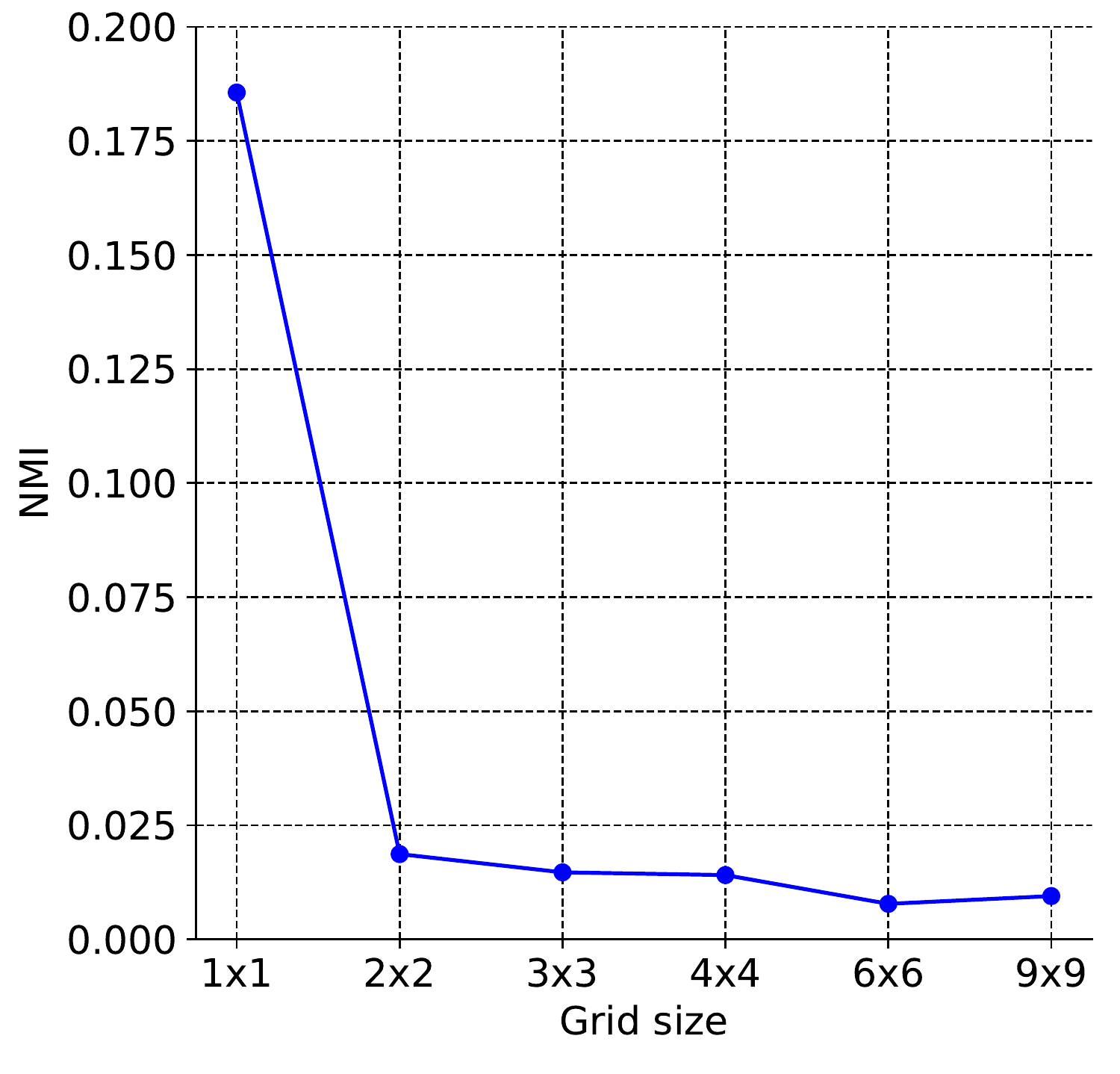}
\label{fig:nmi_analysis-a}}
\vspace{-2pt}
\caption{{\bf Impact of class-destructive transformation}. NMI between the clusters of the features learned on the transformed images at different grid partitions and the ground truth domain labels (a) and class labels (b). SVHN and SynDigits are used.
} 
\label{fig:nmi_analysis}
\vspace{-9pt}
\end{figure}

This idea is supported by a quick analysis. We measure the consistency between the clusters of the features learned on the transformed images (by the self-supervised learning algorithm described in the next paragraph) and the ground truth class/domain labels by normalized mutual information (NMI).
The results are shown in Fig~\ref{fig:nmi_analysis}.
As the number of grid partitions increases, NMI to the domain labels becomes high within a certain range, while that to the class labels sharply decreases.
These results show that by taking an appropriate number of grid partitions, domain-variant features that are invariant to the class information can be learned.


\medskip
\noindent {\bf Algorithm}.
Our algorithm is simple.
We first apply the class-destructive transformation to the training images to obtain the transformed images and then perform self-supervised learning on these images to obtain class-invariant features. Finally, we apply clustering to the learned features to assign a domain label to each cluster.

We use \cite{chen2020simple} for self-supervised learning.
A feature extractor $f$ is learned by minimizing the normalized temperature-scaled cross-entropy loss~\cite{chen2020simple} (with the temperature of $0.5$) between the features of two images augmented from the same image. The two images are generated as $t(t_{s}(\mathbf{x}))$ and $t'(t_{s}(\mathbf{x}))$, where $t$ and $t'$ are two random augmentation operations (e.g., random crop) and $t_s$ is the class-destructive transformation.
After training, we estimate the domain labels for all the images by applying clustering to their features. We use a Gaussian mixture for clustering.

\medskip
\noindent {\bf Discussion}. 
A relevant idea to our approach is ``jigsaw", a popular pretext task for self-supervised learning~\cite{noroozi2016unsupervised}, which has also been applied to UDA and domain generalization~\cite{bucci2019tackling,carlucci2019domain}.
Our work differs from them in both method and purpose.
These methods aim to learn {\em class-dependent} features like object shape and part connections by letting the network solve the task of undoing shuffled images in pixel blocks.
Our method, in contrast, aims to learn {\em class-independent} features and therefore does not solve the jigsaw task. 
Technically, this is a slight difference, but it has the exact opposite effect.
More recently, a self-supervised learning method for learning pretext invariant features has been proposed~\cite{misra2020self}. Although it does not address UDA, our approach is somewhat similar to it.
In this work, we show that the class-dependent information is corrupted more quickly than the domain-dependent information as the pixel block size is decreased, and, based on this finding, we propose a simple self-supervised approach to learning class-independent and domain-sensitive features to solve UDA with unknown domain labels. To the best of our knowledge, these points have never been explored before.
Another recent work uses self-supervised clustering for UniDA~\cite{saito2020universal}. Unlike ours, it assumes that the domain labels are known.

\subsection{Classifier Learning} \label{sec:classifier_learning}
The second step trains our class classifier network. Once the domain labels are estimated, we can train the classifier through domain-adversarial classifier learning.

\medskip
\noindent{\bf Domain-adversarial Classifier Learning}.
We follow a standard domain-adversarial learning approach~\cite{ganin2014unsupervised} to train our classifier network.
The network consists of three major parts: a (shared) feature extractor $G_f$, a class label predictor $F_y$, and a domain classifier $F_d$. 
Our class classifier is defined as $F = G_f \circ F_y$. 
An image $\mathbf{x}$ is fed to $G_f$ to extract the feature and then mapped by $F_y$ to the class label $y$ and by $F_d$ to the domain label $d$. 
With the aim of learning features discriminative to the classes but domain-invariant, the entire network is trained by solving the following problems:
\begin{align}
\begin{aligned}
& \min_{G_f,F_y} \mathcal{L}_{y} - \lambda \mathcal{L}_{d}, \\
& ~\min_{F_d}~~\mathcal{L}_{d},
\end{aligned} \label{eq:model_loss}
\end{align}
where $\mathcal{L}_y$ is a class classification loss and $\mathcal{L}_d$ is a domain classification loss. 
We use softmax cross-entropy for both, with the ground truth class labels and the domain labels estimated by our domain label estimation method.
The two problems can be efficiently minimized at the same time by using the gradient reversal layer (GRL)~\cite{ganin2014unsupervised}.

\medskip
\noindent{\bf Handling Unknown Classes and Unlabeled Samples}.
So far, we considered the case where neither unknown classes nor unlabeled samples of known classes exist. In our new problems, however, both exist.
Let us assume that all the unlabeled samples are of unknown classes, and `UNK' (the class label for unknown classes) is assigned to all.
In reality, some of them are truly of unknown classes, while others are actually of known ones.
We treat this problem as ``learning with noisy labels" and solve it by estimating the true class labels of all the unlabeled samples and optimizing the network parameters simultaneously.

\begin{table*}
    \centering
    \caption{{\bf Results for Digits in GDA1}. The best and second-best are highlighted in bold and underlined, respectively.}
    \label{tab:exp-Digit}
    \vspace{-9pt}
    \scalebox{0.62}{
    \begin{tabular}{l|ccc|ccc|ccc|ccc|ccc|ccc||ccc}
    \hline
         & \multicolumn{3}{c|}{sv(0-3), sy(4-7)} & \multicolumn{3}{c|}{sv(0-3), mt(4-7)} & \multicolumn{3}{c|}{sv(0-2), sy(3-5), mt(6-8)} & \multicolumn{3}{c|}{\begin{tabular}{c}sv(0,1), sy(2,3),\\mt(4,5), mm(6,7)\end{tabular}} & \multicolumn{3}{c|}{sv(0-5), sy(2-7)} & \multicolumn{3}{c||}{sv(0-5), mt(2-7)} & \multicolumn{3}{c}{Avg.}\\
         & OS* & UNK & HOS & OS* & UNK & HOS & OS* & UNK & HOS & OS* & UNK & HOS & OS* & UNK & HOS & OS* & UNK & HOS & OS* & UNK & HOS \\ \hline
        Labeled Only & 69.50  & - & - & 20.82  & - & - & 42.95  & - & - & 42.31  & - & - & 71.32  & - & - & 25.71  & - & - & 45.44  & - & - \\
        MCD~\cite{saito2018maximum} & 84.12  & - & - & 23.17  & - & - & 48.07  & - & - & 41.07  & - & - & 79.61  & - & - & 24.77  & - & - & 50.14  & - & - \\
        OSBP~\cite{saito2018open} & 64.33  & 84.91  & \underline{73.20}  & 7.37  & 79.32  & \underline{13.49}  & 33.95  & 60.84  & \underline{43.58}  & 19.48  & 66.47  & \underline{30.13}  & 33.15  & 70.53  & 45.10  & 10.93  & 80.33  & \underline{19.24}  & 28.20  & 73.73  & \underline{37.46}  \\
        ROS~\cite{bucci2020effectiveness} & 43.62  & 86.98  & 58.10  & 5.10  & 60.23  & 9.40  & 18.15  & 67.07  & 28.57  & 15.36  & 79.40  & 25.74  & 50.90  & 80.37  & \underline{62.33}  & 0.67  & 56.04  & 1.32  & 22.30  & 71.68  & 30.91\\
        UAN~\cite{you2019universal} & 61.50  & 0.00  & 0.00  & 14.50  & 0.70  & 1.34  & 39.96  & 0.92  & 1.80  & 33.89  & 0.00  & 0.00  & 65.72  & 6.55  & 11.91  & 24.90  & 2.21  & 4.06  & 40.08  & 1.73  & 3.18  \\ \hline
        {\bf Ours} & 86.18  & 91.19  & {\bf 88.61}  & 70.50  & 85.31  & {\bf 77.20}  & 79.76  & 84.08  & {\bf 81.86}  & 66.02  & 79.97  & {\bf 72.33}  & 84.83  & 85.75  & {\bf 85.29}  & 73.46  & 76.92  & {\bf 75.15}  & 76.79  & 83.87  & {\bf 80.07}  \\ \hline
    \end{tabular}
    }
    \vspace{-9pt}
\end{table*}


\begin{table}
    \centering
    \caption{{\bf Results for Office-31 in GDA1}. The best and second-best are highlighted in bold and underlined, respectively.}
    \label{tab:exp-Office31}
    \vspace{-9pt}
    \scalebox{0.48}{
    \begin{tabular}{l|ccc|ccc|ccc||ccc}
    \hline
         & \multicolumn{3}{c|}{D(0-9), W(10-19)} &  \multicolumn{3}{c|}{A(0-9), D(10-19)} &  \multicolumn{3}{c||}{W(0-9), A(10-19)} & \multicolumn{3}{c}{Avg.} \\
         & OS* & UNK & HOS & OS* & UNK & HOS & OS* & UNK & HOS & OS* & UNK & HOS \\ \hline
        Labeled Only & 93.25  & - & - & 30.91  & - & - & 41.35  & - & - & 55.17  & - & - \\
        MCD~\cite{saito2018maximum} & 86.15  & - & - & 39.55  & - & - & 23.43  & - & - & 49.71  & - & - \\
        OSBP~\cite{saito2018open} & 95.56  & 87.91  & \underline{91.58}  & 0.42  & 99.91  & 0.84  & 3.56  & 99.84  & 6.87  & 33.18  & 95.89  & 33.10  \\
        ROS~\cite{bucci2020effectiveness} & 93.95  & 91.65  & {\bf 92.79}  & 38.57  & 74.92  & \underline{50.92}  & 36.82  & 74.20  & \underline{49.22}  & 56.45  & 80.26  & \underline{64.31}  \\
        UAN~\cite{you2019universal} & 98.03  & 42.64  & 59.43  & 6.42  & 28.17  & 10.46  & 51.72  & 32.13  & 39.64  & 52.06  & 34.31  & 36.51  \\ \hline
        {\bf Ours} & 91.71  & 82.19  & 86.69  & 76.64  & 75.33  & {\bf 75.98}  & 76.87  & 84.28  & {\bf 80.40}  & 81.74  & 80.60  & {\bf 81.02}  \\ \hline
    \end{tabular}
    }
    \vspace{0mm}
\end{table}

Our approach is based on the idea of joint label-network optimization~\cite{tanaka2018joint}. 
We first train an initial class classifier $F$ by solving the problem~Eq.~\eqref{eq:model_loss} using only the labeled samples and then assign an initial pseudo class label to each unlabeled sample $\mathbf{x}$ using the initial classifier.
Our assumption is that if an unlabeled sample $\mathbf{x}$ belongs to a known class $y \in \mathcal{L}$, the class probability of the corresponding class predicted by the initial classifier is high, and if it belongs to an unknown class $y \in \mathcal{C}\setminus\mathcal{L}$, the probabilities are comparable between the classes.
With this assumption, we determine the initial pseudo class label $y$ of $\mathbf{x}$ based on the entropy of the class probability. Specifically,
\begin{align}
\begin{aligned}
& y = \begin{cases}
    {\rm UNK} & (H(y|\mathbf{x}) > \sigma), \\
    \argmax_{k} F(\mathbf{x})[k] \ & ({\rm otherwise}),\\
  \end{cases}
\end{aligned} \label{eq:entropy_initialization}
\end{align}
where $F(\mathbf{x})[k]$ gives the class probability of the $k$-th class. $H(y|\mathbf{x})$ is the entropy, i.e., $H(y|\mathbf{x}) = \sum_{k}F(\mathbf{x})[k]\log F(\mathbf{x})[k]$, and $\sigma$ is a threshold.
Given the initial pseudo class labels, we update the network parameters on both labeled and unlabeled samples by solving~Eq.~\eqref{eq:model_loss} with the additional regularization term $\mathcal{L}_p$ in~\cite{tanaka2018joint} used to prevent the assignment of all labels to a single class.
Note that we add one extra output node so that the network can directly output the probability of UNK. 
Following~\cite{tanaka2018joint}, we update the estimated class labels for unlabeled samples as $y = {\argmax}_k~F(\mathbf{x})[k]$ during training.

\begin{figure}[t]
\vspace{-13pt}
\centering
\subfloat[][OSBP]{\includegraphics[clip, width=.5\columnwidth]{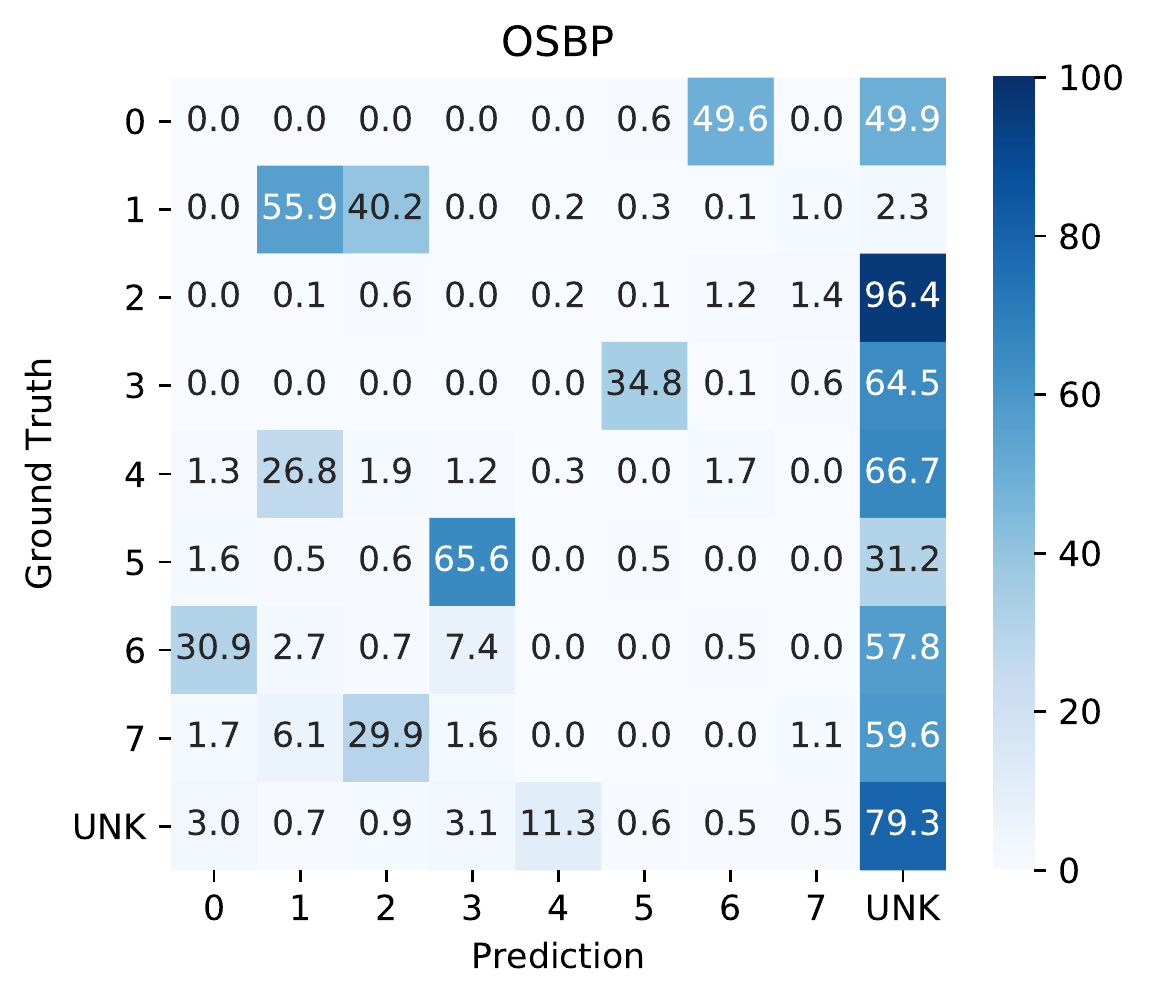}
\label{fig:confusion-matrix-OSBP}}
\subfloat[][Ours]{\includegraphics[clip, width=.5\columnwidth]{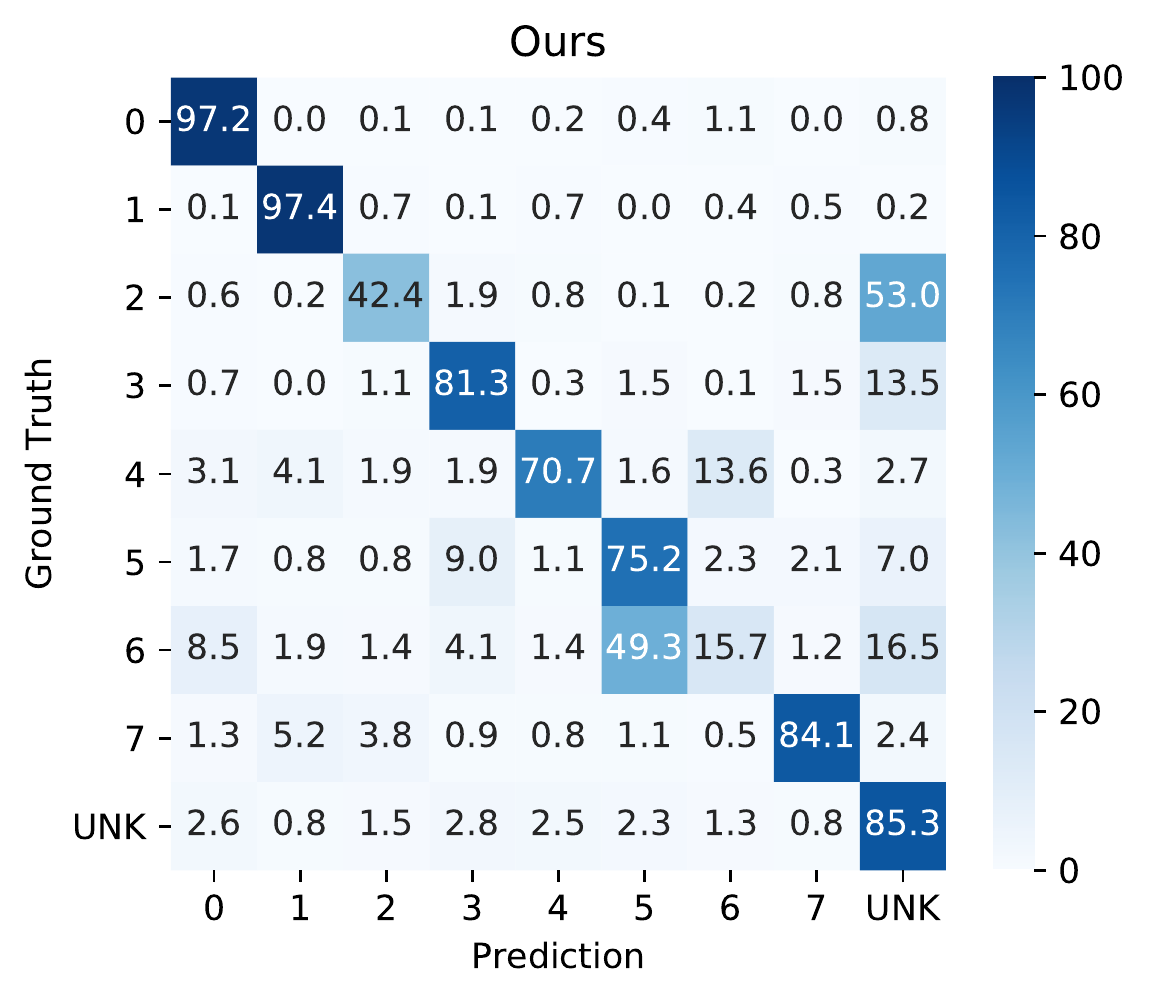}
\label{fig:confusion-matrix-Ours}}
\caption{{\bf Confusion matrices}. The results of OSBP and Ours in ``sv(0-3), mt(4-7)'' are reported.}
\label{fig:confusion-matrix}
\vspace{-9pt}
\end{figure}


\section{Experiments}\label{sec:experiments} 
We first show the results on the new problems, GDA1 and GDA2, in Sec.~\ref{sec:exp-new} and then those on the three existing problems covered by GDA, i.e., OSDA, MS-OSDA and BTDA, in Sec.~\ref{sec:exp-exi}.

\subsection{Experiments on New UDA Problems} \label{sec:exp-new}
We show the results in GDA1 and GDA2. In summary, our method outperforms existing methods. 

\medskip
\noindent {\bf Datasets}.
We use Digits and Office-31.
Digits is composed of four digit datasets: SVHN (sv)~\cite{svhn}, SynDigits (sy)~\cite{ganin2014unsupervised}, MNIST (mt)~\cite{mnist}, and MNIST-M (mm)~\cite{ganin2014unsupervised}. Following \cite{chen2019blending}, we use 25,000 samples in sv, sy, and mt for training and 9,000 samples for testing. For mm, we use 21,661 samples for training and 7,854 samples for testing.
Office-31~\cite{saenko2010adapting} is composed of three domains: 
Amazon (A), DSLR (D), and Webcam (W) having 4,110 samples of $31$ object classes.

\begin{table}
    \centering
    \caption{{\bf Results for Digits in GDA2}. The best and second-best are highlighted in bold and underlined, respectively.} 
    \label{tab:exp-Digit-semi}
    \vspace{-9pt}
    \scalebox{0.49}{
    \begin{tabular}{l|ccc|ccc|ccc||ccc}
    \hline
        & \multicolumn{3}{c|}{sv(0-3), mt(4-7)} &  \multicolumn{3}{c|}{sv(0-2), sy(3-5), mt(6-8)} &  \multicolumn{3}{c||}{\begin{tabular}{c}sv(0,1), sy(2,3),\\mt(4,5), mm(6,7)\end{tabular}} & \multicolumn{3}{c}{Avg.}\\
         & OS* & UNK & HOS & OS* & UNK & HOS & OS* & UNK & HOS & OS* & UNK & HOS \\ \hline
        OSBP~\cite{saito2018open} & 52.45  & 86.32  & \underline{65.25}  & 55.29  & 26.24  & \underline{35.59}  & 41.33  & 79.41  & \underline{54.37}  & 49.69  & 63.99  & \underline{51.74}  \\
        ROS~\cite{bucci2020effectiveness} & 0.20  & 68.95  & 0.40  & 12.45  & 89.34  & 21.85  & 8.15  & 81.15  & 14.81  & 6.93  & 79.81  & 12.36  \\
        UAN~\cite{you2019universal} & 56.96  & 0.17  & 0.34  & 59.86  & 0.96  & 1.89  & 48.99  & 0.00  & 0.00  & 55.27  & 0.38  & 0.74  \\ \hline
        {\bf Ours} & 76.95  & 84.54  & {\bf 80.57}  & 77.26  & 74.51  & {\bf 75.86}  & 69.67  & 78.18  & {\bf 73.68}  & 74.63  & 79.08  & {\bf 76.70}  \\ \hline
    \end{tabular}
    }
    \vspace{0mm}
\end{table}

\begin{table}
    \centering
    \caption{{\bf Results for Office-31 in GDA2}. The best and second-best are highlighted in bold and underlined, respectively.} 
    \label{tab:exp-Office31-semi}
    \vspace{-9pt}
    \scalebox{0.5}{
    \begin{tabular}{l|ccc|ccc|ccc||ccc}
    \hline
         & \multicolumn{3}{c|}{D(0-9), W(10-19)} &  \multicolumn{3}{c|}{A(0-9), D(10-19)} &  \multicolumn{3}{c||}{W(0-9), A(10-19)} & \multicolumn{3}{c}{Avg.} \\
         & OS* & UNK & HOS & OS* & UNK & HOS & OS* & UNK & HOS & OS* & UNK & HOS \\ \hline
        OSBP~\cite{saito2018open} & 93.57  & 92.96  & {\bf 93.26}  & 46.20  & 77.75  & \underline{57.96}  & 51.63  & 78.66  & \underline{62.34}  & 63.80  & 83.12  & \underline{71.19} \\
        ROS~\cite{bucci2020effectiveness} & 92.35  & 90.77  & 91.55  & 40.32  & 71.16  & 51.47  & 27.69  & 76.85  & 40.71  & 53.45  & 79.59  & 61.25  \\
        UAN~\cite{you2019universal} & 97.24  & 46.81  & 63.20  & 65.96  & 23.58  & 34.74  & 65.83  & 28.15  & 39.44  & 76.34  & 32.85  & 45.79  \\ \hline
        {\bf Ours} & 90.33  & 93.41  & \underline{91.84}  & 72.44  & 65.08  & {\bf 68.56}  & 72.99  & 83.89  & {\bf 78.06}  & 78.59  & 80.79  & {\bf 79.49}  \\ \hline
    \end{tabular}
    }
    \vspace{-9pt}
\end{table}

\begin{table*}
    \centering
    \caption{{\bf Results for Office-31 in OSDA}. The best and second-best are highlighted in bold and underlined, respectively.}
    \label{tab:exp_osda_office31}
    \vspace{-9pt}
    \scalebox{0.66}{
    \begin{tabular}{l|ccc|ccc|ccc|ccc|ccc|ccc||ccc}
    \hline
         & \multicolumn{3}{c|}{A$\rightarrow$W} & \multicolumn{3}{c|}{A$\rightarrow$D}& \multicolumn{3}{c|}{D$\rightarrow$W} & \multicolumn{3}{c|}{W$\rightarrow$D} & \multicolumn{3}{c|}{D$\rightarrow$A} & \multicolumn{3}{c||}{W$\rightarrow$A} & \multicolumn{3}{c}{\bf Avg.} \\
         & OS* & UNK & HOS & OS* & UNK & HOS & OS* & UNK & HOS & OS* & UNK & HOS & OS* & UNK & HOS & OS* & UNK & HOS & OS* & UNK & HOS \\ \hline
        STA\textsubscript{sum}~\cite{liu2019separate} & 92.1 & 58.0 & 71.0 & 95.4 & 45.5 & 61.6 & 97.1 & 49.7 & 65.5 & 96.6 & 48.5 & 64.4 & 94.1 & 55.0 & 69.4 & 92.1 & 46.2 & 60.9 & 94.6 & 50.5 & 65.5\\
        STA\textsubscript{max}~\cite{liu2019separate} & 86.7 & 67.6 & 75.9 & 91.0 & 63.9 & 75.0 & 94.1 & 55.5 & 69.8 & 84.9 & 67.8 & 75.2 & 83.1 & 65.9 & 73.2 & 66.2 & 68.0 & 66.1 & 84.3 & 64.8 & 72.5\\
        OSBP~\cite{saito2018open} & 86.8 & 79.2 & \underline{82.7} & 90.5 & 75.5 & {\bf 82.4} & 97.7 & 96.7 & \underline{97.2} & 99.1 & 84.2 & 91.1 & 76.1 & 72.3 & 75.1 & 73.0 & 74.4 & 73.7 & 87.2 & 80.4 & 83.7\\
        UAN~\cite{you2019universal} & 95.5 & 31.0 & 46.8 & 95.6 & 24.4 & 38.9 & 99.8 & 52.5 & 68.8 & 81.5 & 41.4 & 53.0 & 93.5 & 53.4 & 68.0 & 94.1 & 38.8 & 54.9 & 93.4 & 40.3 & 55.1 \\
        ROS~\cite{bucci2020effectiveness} & 88.4 & 76.7 & 82.1 & 87.5 & 77.8 & {\bf 82.4} & 99.3 & 93.0 & 96.0 & 100.0 & 99.4 & {\bf 99.7} & 74.8 & 81.2 & \underline{77.9} & 69.7 & 86.6 & \underline{77.2} & 86.6 & 85.8 & {\bf 85.9}\\ \hline
        {\bf Ours} & 82.5 & 84.0 & {\bf 83.2} & 73.8 & 83.2 & \underline{77.9} & 96.7 & 98.0 & {\bf 97.4} & 99.1 & 96.1 & \underline{97.6} & 67.8 & 92.4 & {\bf 78.2} & 70.8 & 85.1 & {\bf 77.3} & 81.8 & 89.8 & \underline{85.3} \\ \hline
    \end{tabular}
    }
    \vspace{0mm}
\end{table*}

\begin{table*}
    \centering
    \caption{{\bf Results for Office-Home in OSDA}. The best and second-best are highlighted in bold and underlined, respectively.}
    \label{tab:exp_osda_officehome}
    \vspace{-9pt}
    \scalebox{0.66}{
    \begin{tabular}{l ccc ccc ccc ccc ccc ccc ccc}
    \cline{1-19}
         \multicolumn{1}{c|}{} & \multicolumn{3}{c|}{Pr$\rightarrow$Rw} &  \multicolumn{3}{c|}{Pr$\rightarrow$Cl} & \multicolumn{3}{c|}{Pr$\rightarrow$Ar} &  \multicolumn{3}{c|}{Ar$\rightarrow$Pr} & \multicolumn{3}{c|}{Ar$\rightarrow$Rw} & \multicolumn{3}{c}{Ar$\rightarrow$Cl} & \\
         \multicolumn{1}{c|}{} & OS* & UNK & \multicolumn{1}{c|}{HOS} & OS* & UNK &  \multicolumn{1}{c|}{HOS} & OS* & UNK &  \multicolumn{1}{c|}{HOS} & OS* & UNK &  \multicolumn{1}{c|}{HOS} & OS* & UNK &  \multicolumn{1}{c|}{HOS} & OS* & UNK &  \multicolumn{1}{c}{HOS} & &\\ \cline{1-19}
        \multicolumn{1}{l|}{STA\textsubscript{sum}~\cite{liu2019separate}} & 78.1  & 66.3  &  \multicolumn{1}{c|}{69.7}  & 44.7  & 71.5  &  \multicolumn{1}{c|}{\bf 55.0}  & 55.4  & 73.7  &  \multicolumn{1}{c|}{\underline{63.1}}  & 68.7  & 59.7  &  \multicolumn{1}{c|}{63.7}  & 81.1  & 50.5  &  \multicolumn{1}{c|}{62.1}  & 50.8  & 63.4  &  \multicolumn{1}{c}{56.3} & & \\
        \multicolumn{1}{l|}{STA\textsubscript{max}~\cite{liu2019separate}} & 76.2  & 64.3  &  \multicolumn{1}{c|}{69.5}  & 44.2  & 67.1  &  \multicolumn{1}{c|}{53.2}  & 54.2  & 72.4  &  \multicolumn{1}{c|}{61.9}  & 68.0  & 48.4  &  \multicolumn{1}{c|}{54.0}  & 78.6  & 60.4  &  \multicolumn{1}{c|}{68.3}  & 46.0  & 72.3  &  \multicolumn{1}{c}{55.8} & & \\
        \multicolumn{1}{l|}{OSBP~\cite{saito2018open}} & 76.2  & 71.7  &  \multicolumn{1}{c|}{\underline{73.9}}  & 44.5  & 66.3  &  \multicolumn{1}{c|}{53.2}  & 59.1  & 68.1  &  \multicolumn{1}{c|}{\bf 63.2}  & 71.8  & 59.8  &  \multicolumn{1}{c|}{65.2}  & 79.3  & 67.5 &  \multicolumn{1}{c|}{72.9} & 50.2  & 61.1  &  \multicolumn{1}{c}{55.1} & & \\
        \multicolumn{1}{l|}{UAN~\cite{you2019universal}}  & 84.0 & 0.1  &  \multicolumn{1}{c|}{0.2} & 59.1 & 0.0 &  \multicolumn{1}{c|}{0.0} & 73.7 & 0.0 &  \multicolumn{1}{c|}{0.0} & 81.1 & 0.0 &  \multicolumn{1}{c|}{0.0} & 88.2 & 0.1 &  \multicolumn{1}{c|}{0.2} & 62.4 & 0.0 &  \multicolumn{1}{c}{0.0} & &\\
        \multicolumn{1}{l|}{ROS~\cite{bucci2020effectiveness}} & 70.8  & 78.4 &  \multicolumn{1}{c|}{\bf 74.4} & 46.5 & 71.2 &  \multicolumn{1}{c|}{\underline{56.3}} & 57.3 & 64.3 &  \multicolumn{1}{c|}{60.6} & 68.4 & 70.3 &  \multicolumn{1}{c|}{\bf 69.3} & 75.8 & 77.2 &  \multicolumn{1}{c|}{\bf76.5} & 50.6 & 74.1 &  \multicolumn{1}{c}{\bf 60.1} & &\\ \cline{1-19}
        
        \multicolumn{1}{l|}{{\bf Ours}} & 65.0 & 78.1 &  \multicolumn{1}{c|}{70.9} & 49.3 & 71.7 &  \multicolumn{1}{c|}{\bf 58.4} & 49.6 & 78.2 &  \multicolumn{1}{c|}{60.7} & 61.5 & 74.5 &  \multicolumn{1}{c|}{\underline{67.4}} & 69.5 & 80.3 & 
        \multicolumn{1}{c|}{\underline{74.5}} & 50.1 & 74.7 &  \multicolumn{1}{c}{\underline{59.9}} & &\\ \hline
        
        \multicolumn{1}{c|}{                 } & \multicolumn{3}{c|}{Rw$\rightarrow$Ar} &  \multicolumn{3}{c|}{Rw$\rightarrow$Pr} & \multicolumn{3}{c|}{Rw$\rightarrow$Cl} &  \multicolumn{3}{c|}{Cl$\rightarrow$Rw} & \multicolumn{3}{c|}{Cl$\rightarrow$Ar} & \multicolumn{3}{c||}{Cl$\rightarrow$Pr} & \multicolumn{3}{c}{\bf Avg.} \\
        \multicolumn{1}{c|}{                 } & OS* & UNK &  \multicolumn{1}{c|}{HOS} & OS* & UNK &  \multicolumn{1}{c|}{HOS} & OS* & UNK &  \multicolumn{1}{c|}{HOS} & OS* & UNK &  \multicolumn{1}{c|}{HOS} & OS* & UNK &  \multicolumn{1}{c|}{HOS} & OS* & UNK &  \multicolumn{1}{c||}{HOS} & OS* & UNK &  \multicolumn{1}{c}{HOS} \\ \hline 
        \multicolumn{1}{l|}{STA\textsubscript{sum}~\cite{liu2019separate}} & 67.9  & 62.3  &  \multicolumn{1}{c|}{65.0}  & 77.9  & 58.0  &  \multicolumn{1}{c|}{66.4}  & 51.4  & 57.9  &  \multicolumn{1}{c|}{54.2}  & 69.8  & 63.2  &  \multicolumn{1}{c|}{66.3}  & 53.0  & 63.9  &  \multicolumn{1}{c|}{57.9}  & 61.4  & 63.5  &  \multicolumn{1}{c||}{62.5} & 63.4 & 62.6 &  \multicolumn{1}{c}{61.9}\\ 
        \multicolumn{1}{l|}{STA\textsubscript{max}~\cite{liu2019separate}} & 67.5  & 66.7  &  \multicolumn{1}{c|}{\underline{67.1}}  & 77.1  & 55.4  &  \multicolumn{1}{c|}{64.5}  & 49.9  & 61.1  &  \multicolumn{1}{c|}{54.5}  & 67.0  & 66.7  &  \multicolumn{1}{c|}{66.8}  & 51.4  & 65.0  &  \multicolumn{1}{c|}{57.4}  & 61.8  & 59.1  &  \multicolumn{1}{c||}{60.4} & 61.8 & 63.3 &  \multicolumn{1}{c}{61.1} \\
        \multicolumn{1}{l|}{OSBP~\cite{saito2018open}} & 66.1  & 67.3  &  \multicolumn{1}{c|}{66.7}  & 76.3  & 68.6  &  \multicolumn{1}{c|}{72.3}  & 48.0  & 63.0  &  \multicolumn{1}{c|}{54.5}  & 72.0  & 69.2  &  \multicolumn{1}{c|}{\underline{70.6}}  & 59.4  & 70.3  &  \multicolumn{1}{c|}{\bf 64.3} & 67.0  & 62.7  &  \multicolumn{1}{c||}{64.7} & 64.1 & 66.3 &  \multicolumn{1}{c}{64.7}\\
        \multicolumn{1}{l|}{UAN~\cite{you2019universal}} & 77.5 & 0.0  &  \multicolumn{1}{c|}{0.2}  & 85.0 & 0.1  &  \multicolumn{1}{c|}{0.1}  & 66.2 & 0.0  &  \multicolumn{1}{c|}{0.0}  & 80.6 & 0.1  &  \multicolumn{1}{c|}{0.2}  & 70.5 & 0.0  &  \multicolumn{1}{c|}{0.0}  & 74.0 & 0.1  &  \multicolumn{1}{c||}{0.2} & 75.2 & 0.0 &  \multicolumn{1}{c}{0.1} \\
        \multicolumn{1}{l|}{ROS~\cite{bucci2020effectiveness}} & 67.0  & 70.8  &  \multicolumn{1}{c|}{\bf 68.8} & 72.0  & 80.0 &  \multicolumn{1}{c|}{\bf 75.7} & 51.5  & 73.0  &  \multicolumn{1}{c|}{\bf 60.4} & 65.3  & 72.2  &  \multicolumn{1}{c|}{68.6}  & 53.6  & 65.5  &  \multicolumn{1}{c|}{58.9}  & 59.8  & 71.6  &  \multicolumn{1}{c||}{\underline{65.2}} & 61.6 & 72.4 &  \multicolumn{1}{c}{\bf66.2} \\ \hline
        \multicolumn{1}{l|}{\bf Ours} & 54.8 & 81.6 &  \multicolumn{1}{c|}{65.6} & 69.5 & 78.9 &  \multicolumn{1}{c|}{\underline{73.8}} & 53.0 & 72.7 &  \multicolumn{1}{c|}{\underline{61.3}} & 64.3 & 78.9 &  \multicolumn{1}{c|}{\bf 70.8} & 47.1 & 80.8 &  \multicolumn{1}{c|}{\underline{59.5}} & 60.5 & 74.5 &  \multicolumn{1}{c||}{\bf 66.8} & 57.8 & 77.1 &  \multicolumn{1}{c}{\underline{65.8}} \\ \hline
    \end{tabular}
    }
    \vspace{-9pt}
\end{table*}

\begin{table}
    \centering
    \caption{{\bf Results for Office-31 in MS-OSDA}. The best and second-best are highlighted in bold and underlined, respectively.}
    \label{tab:exp_mosda_office31}
    \vspace{-9pt}
    \footnotesize
    \scalebox{0.82}{
    \begin{tabular}{l|cc|cc|cc||cc}
    \hline
    & \multicolumn{2}{c|}{A, D $\rightarrow$ W} & \multicolumn{2}{c|}{A, W $\rightarrow$ D} & \multicolumn{2}{c||}{W, D $\rightarrow$ A} & \multicolumn{2}{c}{Avg.} \\
    & OS* & OS & OS* & OS & OS* & OS & OS* & OS\\ \hline
    OSVM~\cite{scheirer2014open} & 71.2 & 51.2 & 84.9 & 56.2 & 58.2 & 61.4 & 71.4 & 56.3\\
    OSVM+DANN~\cite{ganin2014unsupervised} & 65.0 & 83.3 & 68.0 & 91.9 & 51.2 & 37.5 & 61.4 & 70.9\\
    OSBP~\cite{saito2018open} & 94.0 & 90.0 & 93.0 & 89.0 & \underline{79.0} & 75.0 & \underline{88.7} & 84.7\\
    IOSBP~\cite{fu2019openset} & 91.1 & 88.0 & 87.8 & 87.1 & 75.0 & 74.5 & 84.6 & 83.2 \\
    MOSDANET~\cite{rakshit2020mosda} & {\bf 99.0} & {\bf 98.2} & {\bf 99.4} & {\bf 98.3} & {\bf 81.0} & {\bf 79.3} & {\bf 93.1} & {\bf 91.9}\\ \hline
    {\bf Ours} & \underline{94.7} & \underline{94.1} & \underline{94.1} & \underline{94.5} & 76.1 & \underline{75.6} & 88.3 & \underline{88.1}\\ \hline
    \end{tabular}
    }
    \vspace{-9pt}
\end{table}

We use several different settings of the combinations of domains and splits of labeled, unlabeled (known), and unknown classes. 
An example is ``sv(0-3), sy(4-7)" which means that only the sv and sy domains are used, where \{``0", ``1", ``2", ``3"\} in sv and \{``4", ``5", ``6", ``7"\} in sy are labeled and the rest remain unlabeled; \{``8", ``9"\}, which are not labeled in either domain, are treated as unknown.
For GDA2, we follow the same settings as GDA1 but leave half of the samples in each labeled class unlabeled.

\medskip
\noindent {\bf Evaluation Metrics and Baselines}.
We use three evaluation metrics: OS* and UNK, which are the average accuracy on the known and unknown classes, respectively; and HOS, which is the harmonic mean of OS* and UNK.
Although no existing methods are directly applicable to GDA1 and GDA2, we forcibly applied the following methods as baselines by considering the labeled data as the source domain and the unlabeled data as the target domain: Labeled Only (trained with the labeled data only), MCD~\cite{saito2018maximum}, OSBP~\cite{saito2018open}, ROS~\cite{bucci2020effectiveness}, and UAN~\cite{you2019universal}.

\medskip
\noindent {\bf Implementation Details}.
For domain label estimation, we use a two-layer MLP followed by a four-layer CNN for our feature extractor $f$ (see supplementary material for details).
We use the $3 \times 3$ grid for our class-destructive transformation unless otherwise noted. We use random crop and grayscale for our data augmentation. We also use Gaussian blur and the operations used in~\cite{french2018selfensembling} for Digits.
The network is optimized by using Adam~\cite{kingma2014adam} with the learning rate of $1.0 \times 10^{-3}$ for $300$ epochs on Office-31 and $80$ epochs on Digits with the batch size of $512$.

We use a different classifier for each dataset. For Digits, we use a four-layer CNN for our feature extractor $G_f$ and two-layer MLPs for our class label predictor $F_y$ and domain classifier $F_d$ (see supplementary material for detail).
For Office-31, we use ResNet-50~\cite{he2016deep} pretrained on ImageNet with the top classification layers discarded for $G_f$, one fully connected layer for $F_y$, and a three-layer MLP ($1024 \rightarrow 1024 \rightarrow$ \#domains) for $F_d$.
The network is trained for $1000$ epochs using SGD with the initial learning rate of $1.0 \times 10^{-3}$, a momentum of $0.9$, and a weight decay of $5.0 \times 10^{-4}$. The batch size is $256$ for Digits and $16$ for Office-31. The learning rate is changed to $1.0 \times 10^{-1}$ at the $100$th epoch for Digits. The pseudo class labels are initialized at the $100$th epoch and updated at the $200$th epoch (see Sec.~\ref{sec:classifier_learning}). 
As in~\cite{ganin2016domain}, $\lambda$ in Eq.~\ref{eq:model_loss} is scheduled as $\lambda = \frac{2}{1 + \exp[-\gamma \dot p]} - 1$, where $p$ increases linearly from $0$ to $1$ during training. We set $\gamma = 1000$ for Digits and $\gamma = 10$ for Office-31.
We set $\sigma$ in Eq.~\eqref{eq:entropy_initialization} to the median of the entropy within the mini-batch.
We use the true class distribution as the prior for the regularization term $\mathcal{L}_p$ as done in \cite{tanaka2018joint}. Note that performance degradation is not severe if the true distribution is unknown (see supplementary material).

\medskip
\noindent {\bf Results}.
Tables~\ref{tab:exp-Digit} and \ref{tab:exp-Office31} show the results in GDA1.
First, all the baselines fail to achieve adequate performance in many cases. 
We found noticeable decreases in performance for the domains with large domain gaps.
For example, the HOS values of all the baselines cannot be better than $15$\% in ``sv(0-3), mt(4-7)''. In ``W(0-9), A(10-19)'', even the best-performing method, ROS, achieves only $49$\% in HOS.
These results suggest the difficulty of this new problem.
Our method yields adequate performance in all the cases and outperforms all the baselines in most cases; the HOS values of our method exceed $77$\% in ``sv(0-3), mt(4-7)'' and $80$\% in ``W(0-9), A(10-19)''. 
These results show the exceptional effectiveness of our method in this problem.

Fig.~\ref{fig:confusion-matrix} shows confusion matrices for ``sv(0-3), mt(4-7)'' of OSBP (the best competitor in this case) and Ours.
OSBP tends to misclassify \{``0", ``1", ``2", ``3"\} to \{``4", ``5", ``6", ``7"\}, and vice versa (other baselines also show the same tendency).
This implies that the baselines fail to disentangle the class and domain information, which is the reason for their poor performance.
Our method does not suffer from this problem, which emphasizes its effectiveness.

\begin{table}
    \centering
    \caption{{\bf Results for Office-31 in BTDA}. The best and second-best are highlighted in bold and underlined, respectively.}
    \label{tab:exp_blend_office31}
    \vspace{-9pt}
    \footnotesize
    \scalebox{0.82}{
    \begin{tabular}{l|c|c|c||c}
    \hline
        Models & A $\rightarrow$ D, W & D $\rightarrow$ W, A & W $\rightarrow$ A, D & Avg. \\ \hline
        Labeled only & 68.6  & 70.0  & 66.5 & 68.4 \\
        DAN~\cite{long2015learning} & 78.0  & 64.4  & 66.7 & 69.7 \\
        RTN~\cite{long2016unsupervised} & 84.3  & 67.5  & 64.8 & 72.2 \\
        JAN~\cite{long2017deep} & 84.2  & 74.4  & 72.0 & 76.9 \\
        RevGrad~\cite{ganin2014unsupervised} & 78.2  & 72.2  & 69.8 & 73.4 \\
        AMEAN~\cite{chen2019blending} & {\bf 90.1}  & {\bf 77.0}  & {\bf 73.4} & {\bf 80.2} \\ \hline
        {\bf Ours} & \underline{88.8}  & \underline{74.5}  & \underline{73.2} & \underline{78.8} \\ \hline
    \end{tabular}
    }
    \vspace{-9pt}
\end{table}

Tables~\ref{tab:exp-Digit-semi} and \ref{tab:exp-Office31-semi} show the results for GDA2.
Similar to GDA1, the performance of the baselines is unsatisfactory. Ours outperforms all of them in all the settings except for ``D(0-9), W(10-19)'', which proves that Ours is effective even when the class labels are given to only a subset of the samples in the same class.

\subsection{Experiments on Existing UDA Problems} \label{sec:exp-exi}
We show the results in OSDA, MS-OSDA, and BTDA for the protocols used in~\cite{bucci2020effectiveness, rakshit2020mosda, chen2019blending}, respectively. 
Overall, ours is competitive with the state-of-the-art methods, which confirms that it is applicable even when the type of UDA problem cannot be identified in advance.

\medskip
\noindent {\bf Datasets}.
We use Office-31 and Office-Home. The details of Office-31 are given in Sec.~\ref{sec:exp-new}.
Office-Home~\cite{venkateswara2017deep} contains 15,500 samples of $65$ object classes of four domains: Art (Ar), Clipart (Cl), Product (Pr), and Real-World (Rw).

\begin{figure}[t]
\vspace{-13pt}
\centering
\subfloat[][]{\includegraphics[clip, width=.33\columnwidth]{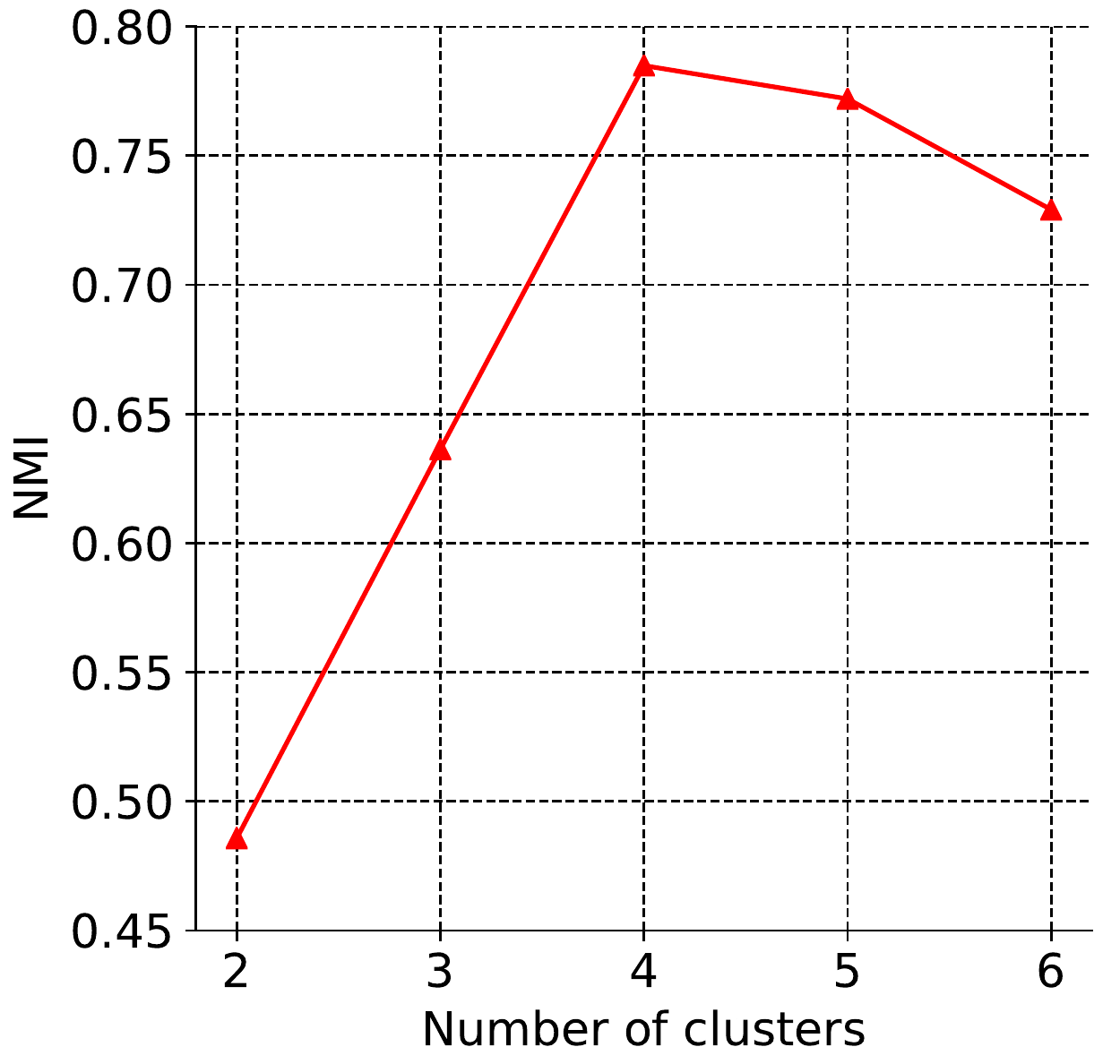}
\label{fig:num_cluster_ablation_nmi}}
\subfloat[][]{\includegraphics[clip, width=.33\columnwidth]{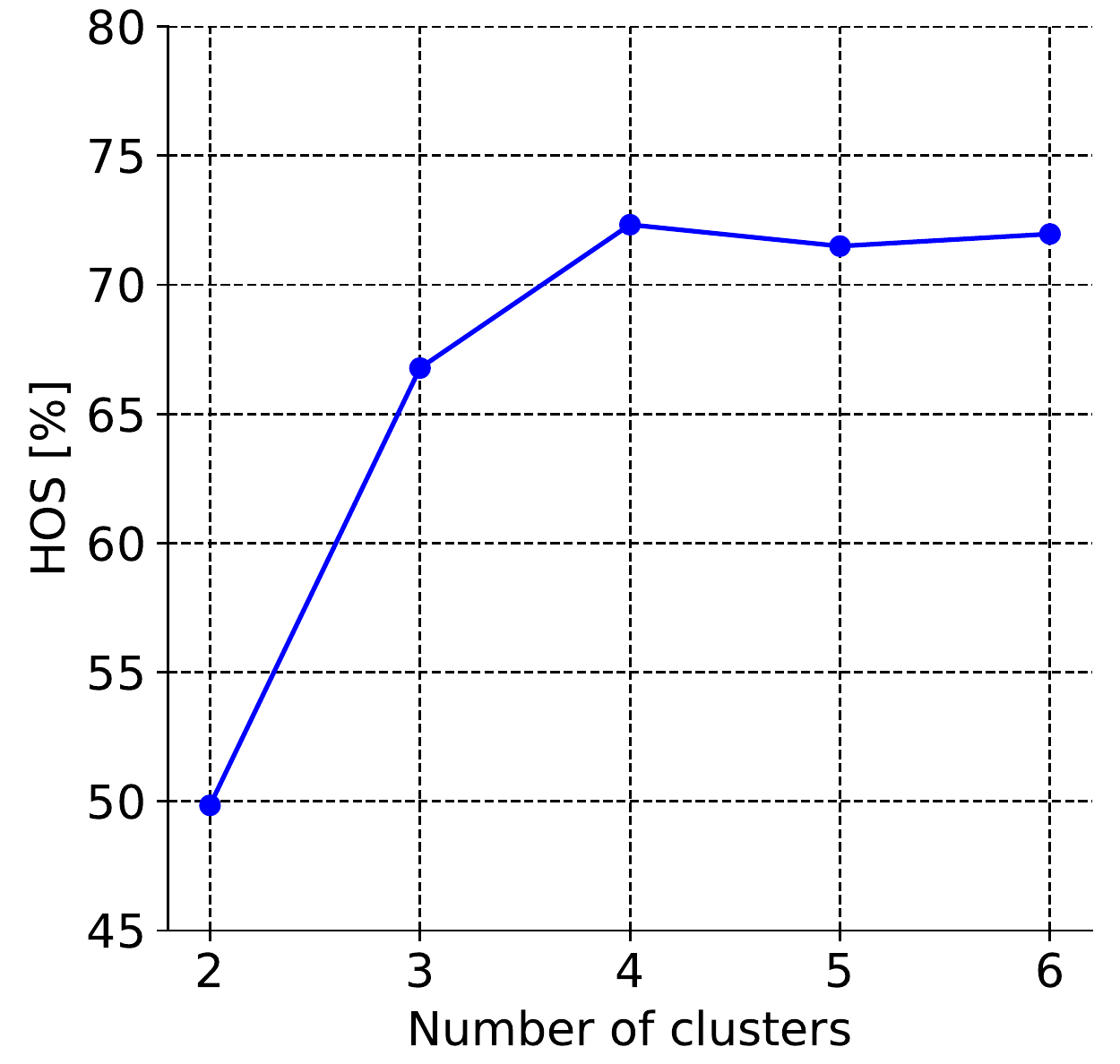}
\label{fig:num_cluster_ablation}}
\subfloat[][]{\includegraphics[clip, width=.33\columnwidth]{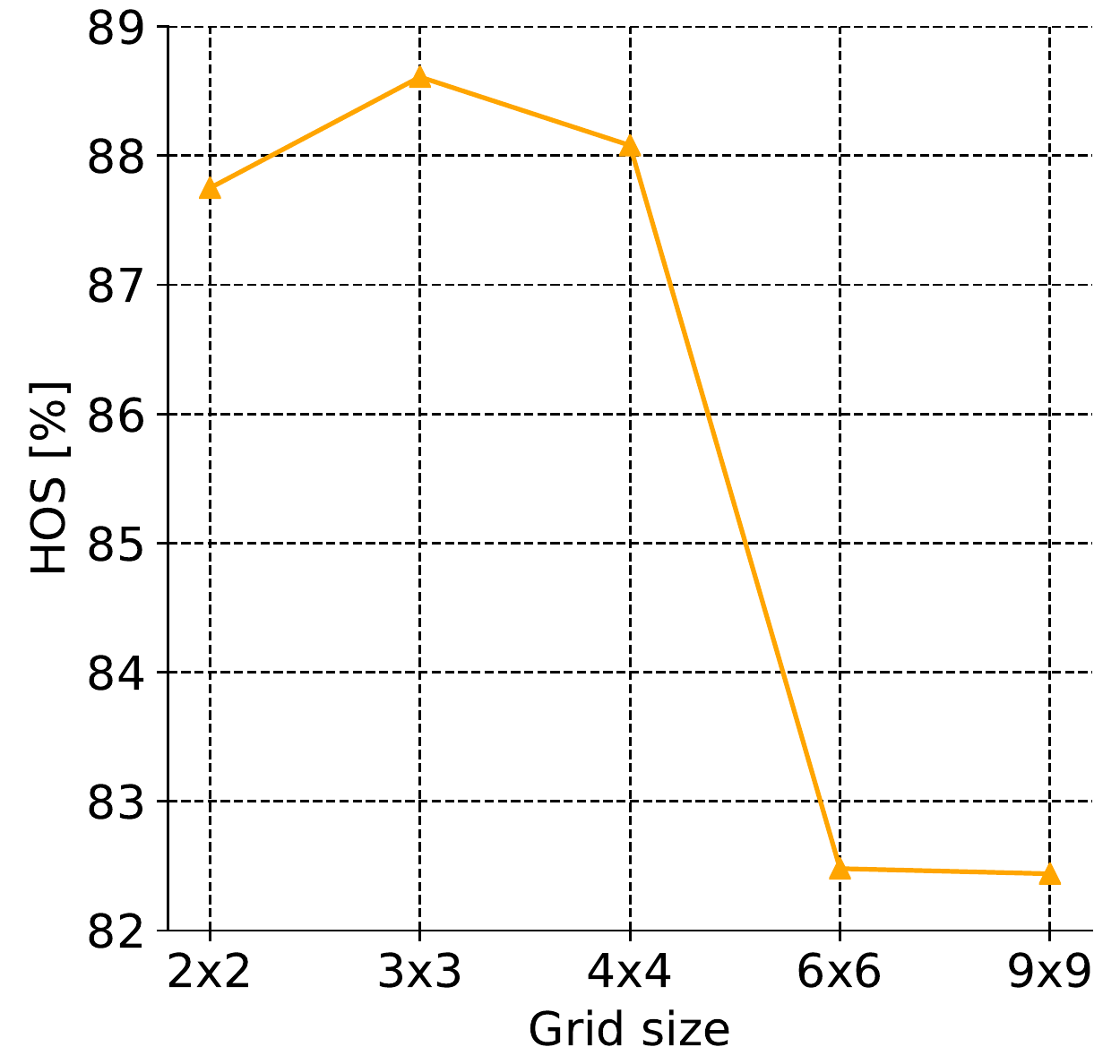}
\label{fig:grid_size_ablation}}
\vspace{-9pt}
\caption{{\bf Sensitivity to hyperparameters}. (a) NMI and (b) HOS vs. the number of clusters in  ``sv(0,1), sy(2,3), mt(4,5), mm(6,7)''. (c) HOS vs. the grid size of class-destractive transformations in ``sv(0-4), sy(4-7)''.} %
\label{fig:analysis}
\vspace{-9pt}
\end{figure}


For OSDA, we follow the protocol used in~\cite{bucci2020effectiveness}. For Office-31, the first $10$ and the last $11$ classes in alphabetical order are used as known and unknown classes, respectively. For Office-Home, the first $25$ and the other $40$ are used as known and unknown classes, respectively. 
We use Office-31 for MS-OSDA~\cite{rakshit2020mosda} and BTDA~\cite{chen2019blending}. For MS-OSDA, we consider all possible combinations of two source domains and one target domain. The first $20$ and the other $11$ classes in alphabetical order are used as shared and open set classes, respectively. For BTDA, we consider all possible combinations of one source domain and two target domains.

\medskip
\noindent {\bf Evaluation Metrics and Baselines}.
We use OS*, UNK, and HOS for OSDA as done in~\cite{bucci2020effectiveness}.
For MS-OSDA, we use OS* and OS~\cite{rakshit2020mosda}; OS is the average accuracy both on known and unknown classes.
For BTDA, we use classification accuracy~\cite{chen2019blending}.
Here, we compare our method with the methods used in the corresponding papers~\cite{bucci2020effectiveness, rakshit2020mosda, chen2019blending}. 

\medskip
\noindent {\bf Implementation Details}.
We use the same networks as in Sec~\ref{sec:exp-new} for both domain label estimation and classification.
We train the classifier for $200$ epochs using SGD with the mini-batch size of $32$, momentum of $0.9$, and weight decay of $5.0 \times 10^{-4}$. The learning rate is initialized to $1.0 \times 10^{-2}$ for $F_y$, $F_d$ and the last layer of $G_f$ and to $1.0 \times 10^{-3}$ for the rest of $G_f$. It is updated at every iteration as $\mathrm{lr} \leftarrow \mathrm{lr}_\mathrm{init} \times (1+0.001 \times i)^{-0.75}$, where $i$ is the number of iterations. The pseudo class labels are initialized at the $40$th epoch and updated at the $80$th epoch.
We use the same scheduling policy for $\lambda$ as in Sec.~\ref{sec:exp-new} but increase $p$ linearly from $0$ to $1$ until the $1000$th iteration with $\gamma = 1.0$.

\medskip
\noindent {\bf Results for OSDA}.
Tables~\ref{tab:exp_osda_office31} and \ref{tab:exp_osda_officehome} show the comparative results for OSDA.
Despite the fact that our method is not designed specifically for OSDA, it is highly competitive with the state-of-the-art methods. 
Compared with ROS, our best competitor, the differences in HOS are only $0.6$\% for Office-31 and $0.4$\% for Office-Home on average.

\medskip
\noindent {\bf Results for MS-OSDA}.
Table~\ref{tab:exp_mosda_office31} shows the results. 
Although Ours is worse than MOSDANET~\cite{rakshit2020mosda}, the state-of-the-art in MS-OSDA, it is the second-best in most cases.

\medskip
\noindent {\bf Results for BTDA}.
Table~\ref{tab:exp_blend_office31} shows the results.
Similar to the results in MS-OSDA, Ours is slightly worse than AMEAN~\cite{chen2019blending} but clearly better than the others.

\subsection{Analysis}
\noindent {\bf Number of Clusters for Domain Label Estimation.}
We evaluate the sensitivity to the number of clusters for domain label estimation. We use ``sv(0,1), sy(2,3), mt(4,5), mm(6,7)'', where the true number of domains is four.
The domain label estimation quality in NMI and HOS are shown in Figs.~\ref{fig:num_cluster_ablation_nmi} and~\ref{fig:num_cluster_ablation}, respectively.
Both reach a maximum at four (the true number of domains), and the HOS value is stable for higher numbers, suggesting that it is not severely sensitive to the number of clusters.

\begin{figure}[t]
\vspace{-9pt}
\centering
\includegraphics[width=\linewidth]{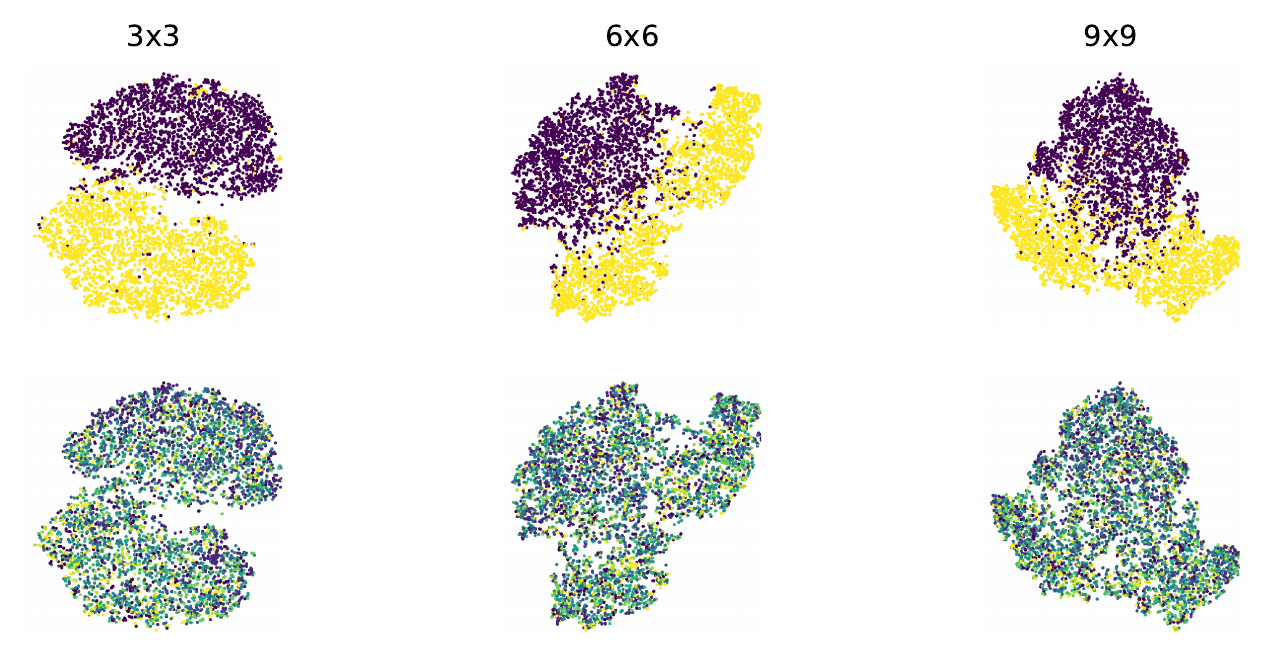}
\caption{{\bf t-SNE visualization}. The features learned by our domain label estimation method in ``sv(0-4), sy(4-7)'' are visualized. The points are color-coded by their domain labels (top) and class labels (bottom).}
\label{fig:grid_plot}
\vspace{-9pt}
\end{figure}

\medskip
\noindent {\bf Grid Size for Class-destructive Transformation}.
Fig.~\ref{fig:grid_size_ablation} reports the performance of our method for various grid sizes in ``sv(0-4), sy(4-7)''.
The accuracy is satisfactory when the partition is coarse but decreases as it becomes finer.
This is not surprising because the domain information is destroyed by over-partitioning, as discussed in Sec.~\ref{sec:domain_label_estimation}.
We show a t-SNE visualization of the features learned by our domain label estimation in Fig~\ref{fig:grid_plot}, which supports the results.

\section{Conclusion}
We proposed Generalized Domain Adaptation (GDA), a general representation of UDA that covers the major UDA variants as well as new challenging settings where existing UDA methods fail.
To solve the new settings, we proposed a novel self-supervised class-destructive learning approach, which estimates domain labels without using any supervision.
Experiments demonstrated that our method outperforms the state-of-the-art methods in the new settings and that it is highly competitive on existing UDA variants.

{\small
\bibliographystyle{ieee_fullname}
\bibliography{egbib}
}

\end{document}


\title{Supplementary Material for \\ Generalized Domain Adaptation}

\author{Yu Mitsuzumi~~~Go Irie~~~Daiki Ikami~~~Takashi Shibata\\
NTT Communication Science Laboratories, NTT Corporation, Japan\\
{\tt\small \{yu.mitsuzumi.ae, daiki.ikami.ef\}@hco.ntt.co.jp, \{goirie, t.shibata\}@ieee.org}
}

\maketitle

\begin{abstract}
In this supplementary material for Generalized Domain Adaptation (main paper), we provide details of the network architectures used in our experiments and present additional experimental results.
\end{abstract}

\section{Network Architectures} 
This section provides the details of the network architectures used in our experiments (Sec.~5 in the main paper). 
We consistently use the same network architecture detailed in Table~\ref{tab: ssl-network} for our domain label estimation throughout all the experiments in the main paper.
The architecture of our classifier network for Digits is shown in Table~\ref{tab:digit-net}.

\section{Additional Results for Office-Home} 
We report additional results for Office-Home in the MS-OSDA and BTDA problems. 
We follow the protocols used in~\cite{rakshit2020mosda, chen2019blending}, as we did in the experiments (Sec.~5.2) described in the main paper.
For MS-OSDA, we consider all four possible combinations of three source domains and one target domain with $45$ shared and $20$ open set classes. 
For BTDA, we consider all four possible combinations of one source domain and three target domains.

\medskip
\noindent {\bf Results for MS-OSDA}.
Table~\ref{tab:exp_mosda_officehome} shows the results. 
Our method is the second best and outperforms MOSDANET~\cite{rakshit2020mosda}, the state-of-the-art MS-OSDA method, in one condition.

\medskip
\noindent {\bf Results for BTDA}.
Table~\ref{tab:exp_blend_officehome} shows the results.
Our method is competitive with AMEAN~\cite{chen2019blending} and consistently better than the other methods.


\begin{table}[t]
    \centering
    \caption{{\bf Network architecture for domain label estimation}.}
    \label{tab: ssl-network}
    \vspace{-9pt}
    \resizebox{\linewidth}{!}{
    \begin{tabular}{c} \hline
          $32 \times 32 \times 1~\mbox{(Digits)}~\mbox{or}~32 \times 32 \times 3~\mbox{(Office-31/Office-Home)}$ Input \\ \hline
          $3 \times 3$ conv. 16 channels, ReLU, BatchNorm \\
          $3 \times 3$ conv. 32 channels, ReLU, BatchNorm \\
          $2 \times 2$ Max Pooling \\
          $3 \times 3$ conv. 64 channels, ReLU, BatchNorm \\
          $3 \times 3$ conv. 64 channels, ReLU, BatchNorm \\
          $2 \times 2$ Max Pooling \\
          Average Pooling \\
          Fully Connected Layer 64, ReLU, BatchNorm \\
          Fully Connected Layer 64 \\ \hline
    \end{tabular}
    }
\end{table}

\begin{table}[t]
    \centering
    \caption{{\bf Network architecture of class classifier for Digits.}}
    \label{tab:digit-net}
    \vspace{-9pt}
    \resizebox{\linewidth}{!}{
    \begin{tabular}{c|c} \hline
        \multicolumn{2}{c}{Feature Extractor ($G_f$)} \\ \hline
        \multicolumn{2}{c}{$32 \times 32 \times 3$ Input} \\
        \multicolumn{2}{c}{InstanceNorm} \\
        \multicolumn{2}{c}{$5 \times 5$ conv. 64 channels, LeakyReLU, SwitchNorm~\cite{SwitchableNorm}} \\
        \multicolumn{2}{c}{$5 \times 5$ conv. 64 channels, LeakyReLU, SwitchNorm~\cite{SwitchableNorm}} \\
        \multicolumn{2}{c}{$3 \times 3$ conv. 128 channels, stride 2, LeakyReLU, SwitchNorm~\cite{SwitchableNorm}} \\
        \multicolumn{2}{c}{$3 \times 3$ conv. 128 channels, stride 2, LeakyReLU, SwitchNorm~\cite{SwitchableNorm}} \\
        \multicolumn{2}{c}{Dropout} \\ \hline
        Class Label Predictor ($F_y$) & Domain Classifier ($F_d$) \\ \hline
                         & Gradient Reversal Layer \\
        Fully Connected Layer 100,  & Fully Connected Layer 100, \\
        ReLU, SwitchNorm~\cite{SwitchableNorm} & ReLU, SwitchNorm~\cite{SwitchableNorm} \\
                & Dropout \\
        Fully Connected Layer 100, & Fully Connected Layer 100,\\
        ReLU, SwitchNorm~\cite{SwitchableNorm} & ReLU, SwitchNorm~\cite{SwitchableNorm} \\
                & Dropout \\
        Fully Connected Layer \#class & Fully Connected Layer \#domain \\
        Softmax & Softmax \\ \hline
    \end{tabular}
    }
\end{table}


\section{Comparison with OCDA}
Liu et al.~\cite{liu2020open} recently proposed a new UDA variant called open compound domain adaptation (OCDA). The OCDA problem deals with adaptation to ``open domains," i.e., domains not present in the training data, and they proposed a solution for the problem.
Although the OCDA problem is not within the scope of our GDA, evaluating the performance of the OCDA method on our GDA problem and the performance of our GDA method on the OCDA problem would be interesting.

\medskip
\noindent {\bf Evaluation of OCDA Method in GDA Problem}. 
We first show the performance of the OCDA method~\cite{liu2020open} in our GDA1 problem. The protocol is exactly the same as the one mentioned in the main paper. Note that we use only the OS* metric because the OCDA method has no mechanism to detect unknown classes.
Table~\ref{tab:ocda} shows the results. Compared with the best baseline in GDA1, OSBP, the OCDA method won some and lost some. Ours is clearly better than the OCDA method in all the setups, which stresses the merit of our method.

\medskip
\noindent {\bf Evaluation of GDA method in OCDA Problem}.
We next report the performance of our method in the OCDA problem. In the experiment, we use SVHN~\cite{svhn} as a source domain, MNIST~\cite{mnist}, MNIST-M~\cite{ganin2014unsupervised}, and USPS~\cite{usps} as compound domains, which consist of multiple target domains without their domain labels, and SynDigits~\cite{ganin2014unsupervised} as an open domain. We compare our method with the OCDA method~\cite{liu2020open} and AMEAN~\cite{chen2019blending}.
The results are shown in Table~\ref{tab:ours_on_ocda}. Note that the symbol $\ddagger$ means that open domain images are treated as being included in compound domains during training.
While Ours is highly competitive with AMEAN, the best baseline used in \cite{liu2020open}, it cannot outperform the OCDA method~\cite{liu2020open}. 
Extending our method to a form applicable to the OCDA problem would be an interesting future direction.


\begin{table}[t]
    \centering
    \caption{{\bf Results for Office-Home in MS-OSDA}. OS values are listed in the table. The best and second-best are highlighted in bold and underlined, respectively.}
    \label{tab:exp_mosda_officehome}
    \vspace{-9pt}
    \footnotesize
    \resizebox{\linewidth}{!}{
    \begin{tabular}{l|c|c|c|c||c} \hline
         & Ar, Cl, Pr  & Ar, Pr, Rw & Pr, Cl, Rw & Ar, Cl, Rr & \multirow{2}{*}{Avg.} \\
         & $\rightarrow$ Rw & $\rightarrow$ Cl & $\rightarrow$ Ar & $\rightarrow$ Pr &  \\ \hline
        OSVM~\cite{scheirer2014open} & 60.2 & 46.3 & 48.6 & 57.0 & 53.0 \\
        OSVM+DANN~\cite{ganin2014unsupervised} & 54.5 & 31.6 & 40.9 & 53.8 & 45.2 \\
        OSBP~\cite{saito2018open} & 53.6 & 38.0 & 46.9 & 54.9 & 48.3 \\
        IOSBP~\cite{fu2019openset} & 64.5 & 46.2 & 54.9 & 66.4 & 58.1 \\
        MOSDANET~\cite{rakshit2020mosda} & {\bf 80.3} & {\bf 67.5} & \underline{60.6} & {\bf 80.0} & {\bf 72.1}\\ \hline
        {\bf Ours} & \underline{78.6} & \underline{59.1} & {\bf 64.8} & \underline{75.5} & \underline{69.5}\\ \hline
    \end{tabular}
    }
\end{table}

\begin{table}[t]
    \centering
    \caption{{\bf Results for Office-Home in BTDA}. Classification accuracy is shown in the table. The best and second-best are highlighted in bold and underlined, respectively.}
    \label{tab:exp_blend_officehome}
    \vspace{-9pt}
    \footnotesize
    \resizebox{\linewidth}{!}{
    \begin{tabular}{l|c|c|c|c||c} \hline
         & Ar $\rightarrow$ & Cl $\rightarrow$ & Pr $\rightarrow$ & Rw $\rightarrow$ & \multirow{2}{*}{Avg.}\\
         & Cl, Pr, Rw & Ar, Pr, Rw & Ar, Cl, Rw & Ar, Cl, Pr & \\ \hline
        Labeled only & 47.6 & 42.6 & 44.2 & 51.3 & 46.4\\
        DAN~\cite{long2015learning} & 55.6 & 56.6 & 48.5 & 56.7 & 54.4\\
        RTN~\cite{long2016unsupervised} & 53.9 & 56.7 & 47.3 & 51.6 & 52.4\\
        JAN~\cite{long2017deep} & 58.3 & 60.5 & 52.2 & 57.5 & 57.1\\
        RevGrad~\cite{ganin2014unsupervised} & 58.4 & 58.1 & 52.9 & 62.1 & 57.9\\
        AMEAN~\cite{chen2019blending} & {\bf 64.3} & {\bf 65.5} & {\bf 59.5} & {\bf 66.7} & {\bf 64.0} \\ \hline
        {\bf Ours} & \underline{63.3} & \underline{63.6} & \underline{58.6} & \underline{64.8} & \underline{62.6} \\ \hline
    \end{tabular}
    }
\end{table}


\section{Additional Analysis}
\noindent {\bf Ablation of Regularization Term $\mathcal{L}_p$~\cite{tanaka2018joint}}.
The regularization term $\mathcal{L}_p$ \cite{tanaka2018joint} requires the true class distribution of the data in advance, which cannot be known in practice.
We evaluated our method without $\mathcal{L}_p$ to ascertain the performance when the distribution is unknown. 
The HOS values on Office-31 were $81.02$ with $\mathcal{L}_p$ vs. $80.34$ without $\mathcal{L}_p$. The gap was only $0.7$\%, which proves the strong robustness of our method.

\begin{figure}[t]
\vspace{-9pt}
\centering
\subfloat[][]{\includegraphics[clip, width=.5\columnwidth]{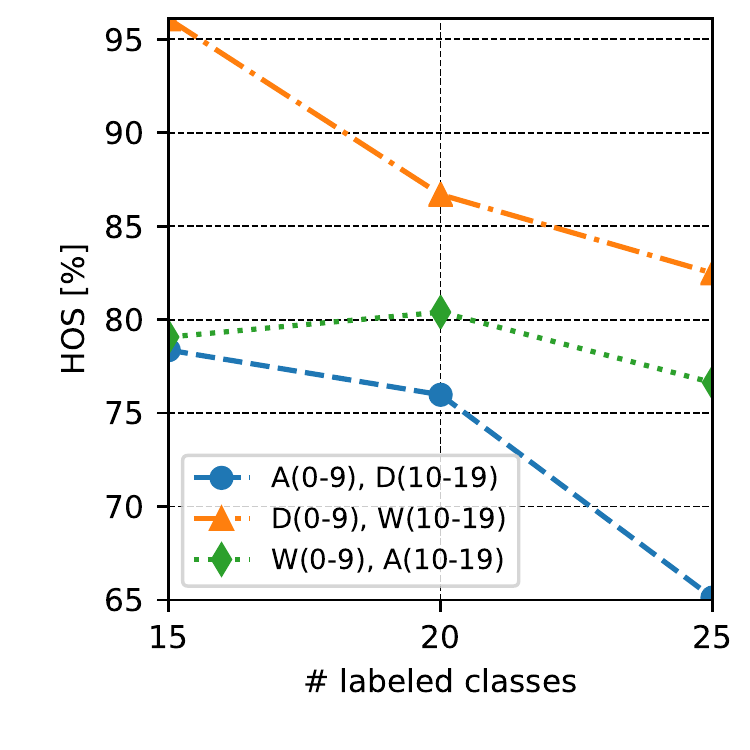}
\label{fig:GDA1_diff}}
\subfloat[][]{\includegraphics[clip, width=.5\columnwidth]{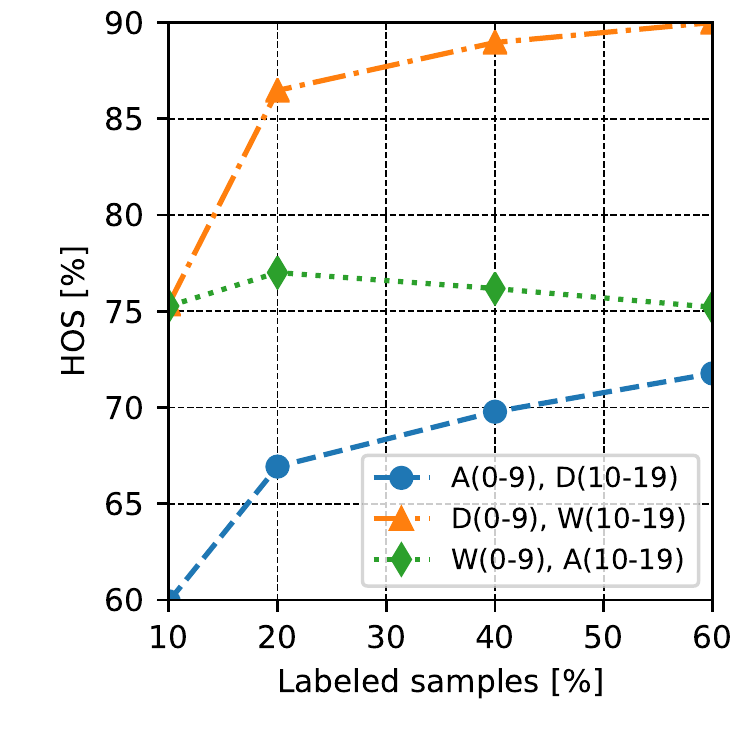}
\label{fig:GDA2_ratio}}
\vspace{-9pt}
\caption{{\bf Performance of GDA with various difficulty levels.}
}
\label{fig:various_GDA}
\end{figure}

\medskip
\noindent {\bf Performance to Different Difficulty Levels}. %
The difficulty of the GDA task varies depending on the number of classes to be classified and the percentage of labeled data.
We evaluate the performance of the proposed method under various difficulty levels of GDA.
Fig.~\ref{fig:GDA1_diff} shows the accuracy for different numbers of labeled classes on Office-31 in GDA1.
As with general multiclass classification, the accuracy decreases as the number increases. 
Fig.~\ref{fig:GDA2_ratio} shows the accuracy for different ratios of labeled samples on Office-31 in GDA2. 
As the ratio decreases, the accuracy decreases.


\begin{table}[t]
\begin{center}
    \caption{{\bf OCDA results for Digits in GDA1.} OS* values are listed in the table. The best are highlighted in bold.}
    \label{tab:ocda}
    \vspace{-9pt}
    \footnotesize
    \scalebox{1.0}{
    \begin{tabular}{l|wc{18mm}|wc{18mm}|wc{18mm}} \hline
    \multirow{2}{*}{Setup}      & sv(0-3), & sv(0-3), & sv(0-2), sy(3-5), \\
                                & sy(4-7) & mt(4-7) & mt(6-8) \\ \hline
      OSBP~\cite{saito2018open} & 64.33 & 7.37 & 33.95 \\
      OCDA~\cite{liu2020open}   & 34.35 & 0.83 & 38.49 \\ 
      Ours                      & {\bf 86.18} & {\bf 70.50} & {\bf 79.76} \\ \hline \hline
    \multirow{2}{*}{Setup}      & sv(0,1), sy(2,3), & sv(0-5), & sv(0-5), \\
                                & mt(4,5), mm(6,7) & mt(2-7)  & mt(2-7) \\ \hline
      OSBP~\cite{saito2018open} & 19.48 & 33.15 & 10.93 \\
      OCDA~\cite{liu2020open}   & 17.08 & 57.22 & 4.58 \\ 
      Ours                      & {\bf 66.02} & {\bf 85.83} & {\bf 73.46} \\ \hline
    \end{tabular}
    }
    \vspace{-9pt}
\end{center}
\end{table}

\begin{table}[t]
    \centering
    \caption{{\bf Results for OCDA problem}. Classification accuracy is reported in the table. The best are highlighted in bold.} 
    \label{tab:ours_on_ocda}
    \vspace{-9pt}
    \footnotesize
    \begin{tabular}{l|ccc|c||c} \hline
                                      Source  & \multicolumn{3}{c|}{Compound Domains} & Open & \\
                                      SVHN    & MNIST & MNIST-M & USPS & SynDigits & Avg. \\ \hline
        AMEAN$\ddagger$~\cite{chen2019blending} & 85.2  & {\bf 65.7}    & 74.3 & 84.4 & 77.4  \\
        OCDA~\cite{liu2020open}               & {\bf 90.9}  & {\bf 65.7}    & {\bf 83.4} & 88.2 & {\bf 82.1}   \\ 
        Ours$\ddagger$                         & 81.7  & 57.9    & 77.4 & {\bf 92.3} & 77.3  \\
        Ours                                  & 81.3  & 59.5    & 76.3 & 87.6       & 76.2  \\ \hline
    \end{tabular}
\end{table}

{\small
\bibliographystyle{ieee_fullname}
\bibliography{egbib}
}